\documentclass[10pt,twocolumn,letterpaper]{article}

\usepackage{iccv}
\usepackage{times}
\usepackage{epsfig}
\usepackage{graphicx}
\usepackage{amsmath}
\usepackage{amssymb}

\usepackage[breaklinks=true,bookmarks=false]{hyperref}

\usepackage{amsmath}
\usepackage{subcaption}

\newcommand\rsx[1]{\left.{#1}\vphantom{\Big|}\right|}

\iccvfinalcopy 

\ificcvfinal\fi

\begin{document}

\title{Light Pollution Reduction in Nighttime Photography}

\author{Chang Liu\\
Shanghai Jiao Tong University\\
\and
Xiaolin Wu\\
McMaster University\\
}

\maketitle
\ificcvfinal\thispagestyle{empty}\fi

\begin{abstract}
   Nighttime photographers are often troubled by light pollution of unwanted artificial lights.  Artificial lights, after scattered by aerosols in the atmosphere, can inundate the starlight and degrade the quality of nighttime images, by reducing contrast and dynamic range and causing hazes.  In this paper we develop a physically-based light pollution reduction (LPR) algorithm that can substantially alleviate the aforementioned degradations of perceptual quality and restore the pristine state of night sky.  The key to the success of the proposed LPR algorithm is an inverse method to estimate the spatial radiance distribution and spectral signature of ground artificial lights. Extensive experiments are carried out to evaluate the efficacy and limitations of the LPR algorithm.
\end{abstract}

\section{Introduction}

A side effect of urbanization is wide spread of nighttime light pollution caused by pervasive artificial lighting and increased density of aerosols in the atmosphere. As light pollution distorts the energy level and spectral signature of natural light in the night, it degrades the quality of nighttime images.  For example, nowadays it is becoming increasingly difficult to capture the Milky Way with a camera; enthusiastic night sky photographers are known to go great distances just to escape the
city lights.  But not everyone has the means and time to travel to a location free of artificial lighting.  Even a weak level of light pollution can ruin artistic appeal of night sky photos, because long exposure required to capture distant faint stars will also accumulate the small amount of artificial lighting to a noticeable level of greyish/brownish background.
In addition, light pollution may be a hindrance to nighttime photography of city scenes as well.
For example, a desired image composition requires shooting far away illuminated buildings or other structures at a spot where nearby street lighting cannot be escaped.

\begin{figure}
  \centering

  \begin{subfigure}{0.47\linewidth}
    \centering
    \includegraphics[width=\linewidth,height=0.66\linewidth]{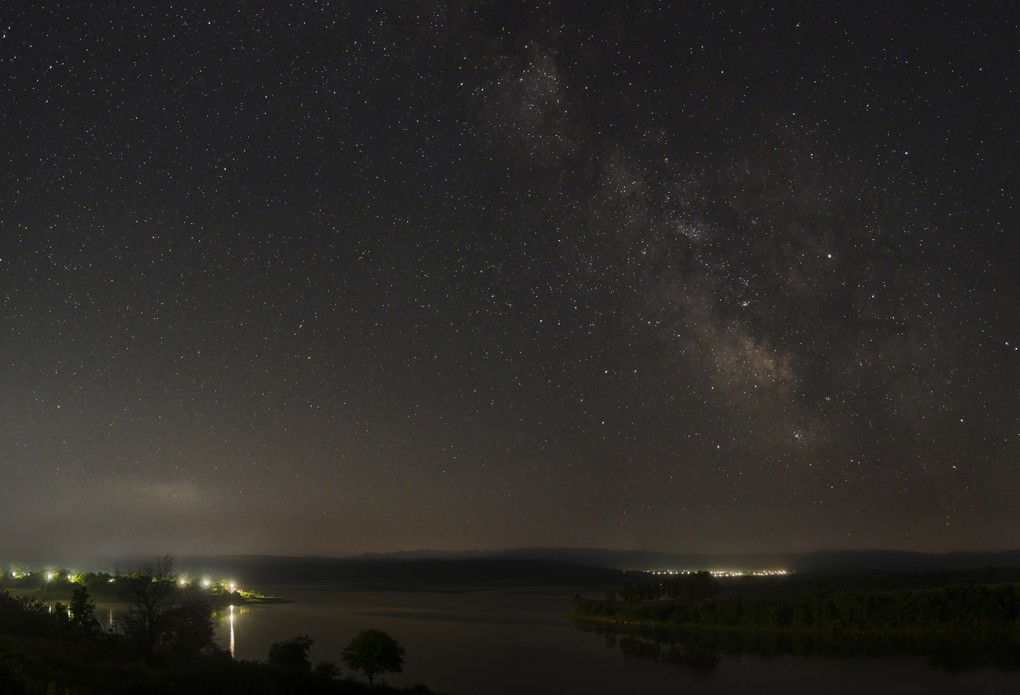}
  \end{subfigure}
  \begin{subfigure}{0.47\linewidth}
     \centering
     \includegraphics[width=\linewidth,height=0.66\linewidth]{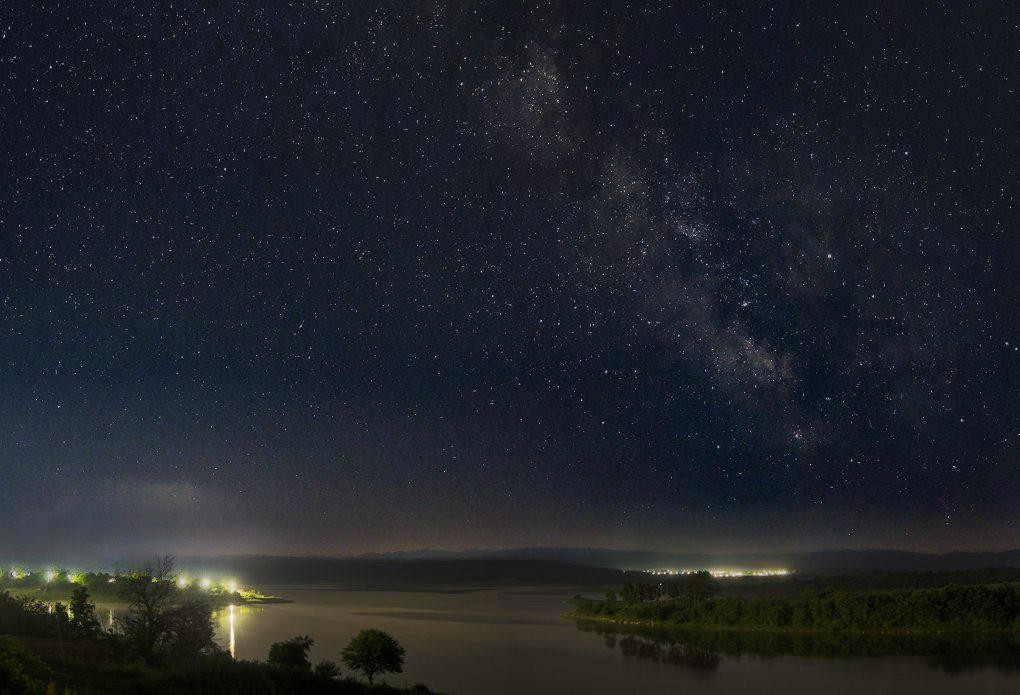}
   \end{subfigure}

  \begin{subfigure}{0.47\linewidth}
    \centering
    \includegraphics[width=\linewidth,height=0.66\linewidth]{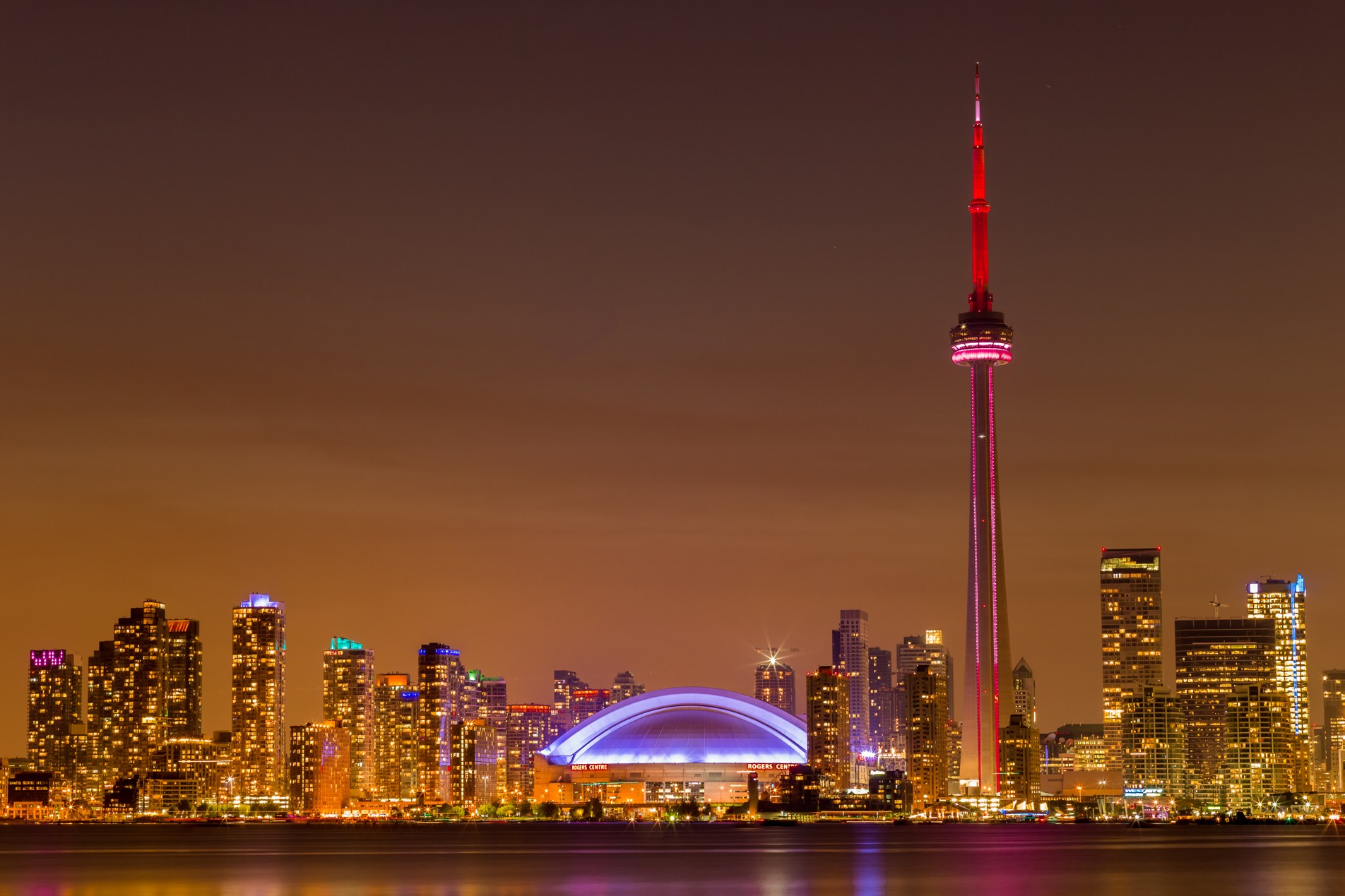}
  \end{subfigure}
  \begin{subfigure}{0.47\linewidth}
      \centering
      \includegraphics[width=\linewidth,height=0.66\linewidth]{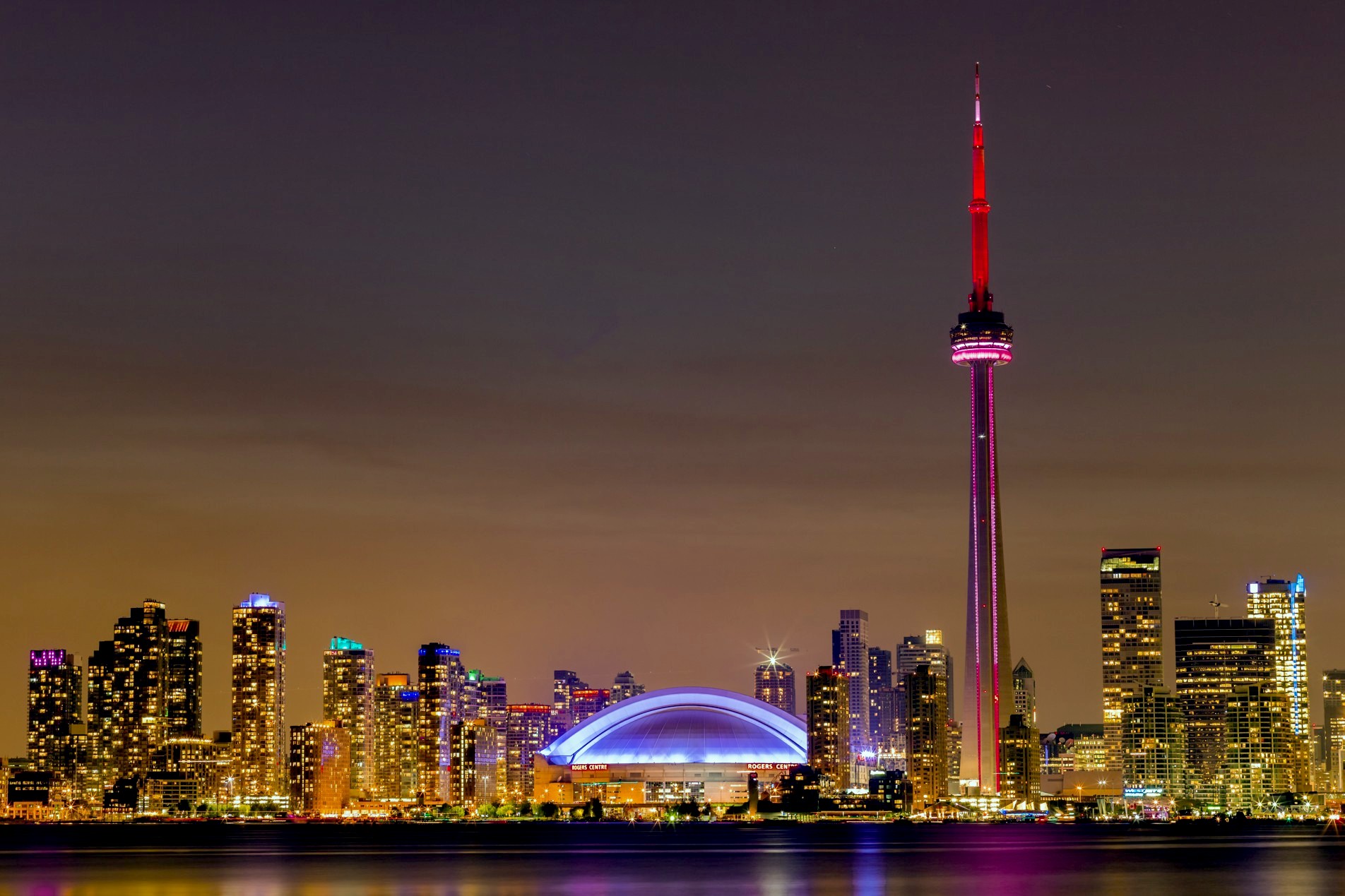}
  \end{subfigure}
  \caption{First column: light-polluted images.  Second column: restored images by the proposed LPR algorithm.}
  \label{fig:title}
  \vspace{-0.5cm}
\end{figure}

As light pollution problem cannot be physically corrected, the only solution is to algorithmically neutralize unwanted effects of light pollution on nighttime photos.  This requires to model the image formation process $\hat{I} = F(I, J)$, where
$I$ is the ideal image free of interference of artificial lighting $J$, and
$\hat{I}$ is the image acquired in presence of $J$, and solve the inverse problem of recovering $I$ from $\hat{I}$.   The above stated modeling and algorithmic problem of removing light pollution in nighttime photography is the main theme and contribution of this paper.  We succeed in designing the algorithm and achieving our design goal as can be previewed in Fig.~\ref{fig:title}.  The ability to image nighttime beauty of pristine nature or sophisticatedly-lit man-made structures is much desired in many existing and potential applications, such as visual arts, high dynamic range imaging, environment study, and astronomy.  To the best of our knowledge, we are the first to attack the problem of light pollution reduction (LPR) for nighttime photography.

Some previous publications on the subject of light pollution are about its adverse effects on the astronomical
observations \cite{garstang1991light,riegel1973light}.  Other papers discuss about the impact of light pollution on human health and environment \cite{chepesiuk2009missing,falchi2011limiting,gaston2015biological}.  In the field of computer graphics, Jensen {\it et al.} studied the problem of realistically rendering night sky images \cite{jensen2001physically}.  Their work is based on physically modeling nighttime illumination effects of astronomical bodies, assuming zero artificial lighting.

In the perspective of image restoration, most relevant to this work is the subject of image dehazing, which has been extensively researched,
including traditional image processing algorithms \cite{fattal2014dehazing,he2010single}, deep learning based algorithms \cite{liu2019griddehazenet,shao2020domain, zhang2018densely}, and some algorithms especially for nighttime dehazing \cite{li2015nighttime,yan2020nighttime}.  The task of light pollution reduction differs from dehazing in two aspects.  Firstly, the degree of light pollution is spatially nonuniform, depending on the geographical distribution and varying strength of artificial lights, and also on how the energy of artificial lighting attenuates in altitude.  The mechanism of light scattering in hazy weather is simpler to model as the sun light can be considered of uniform strength in atmosphere and having a white spectrum.  Secondly, the original signal strength in nighttime images is much weaker than in day time images.  The low signal-to-noise ratio makes the restoration task more difficult in the former case than in the latter case.

\section{Problem background}

\begin{figure}[t]
  \centering
  \setlength{\abovecaptionskip}{1pt}
  \includegraphics[width=0.75\linewidth]{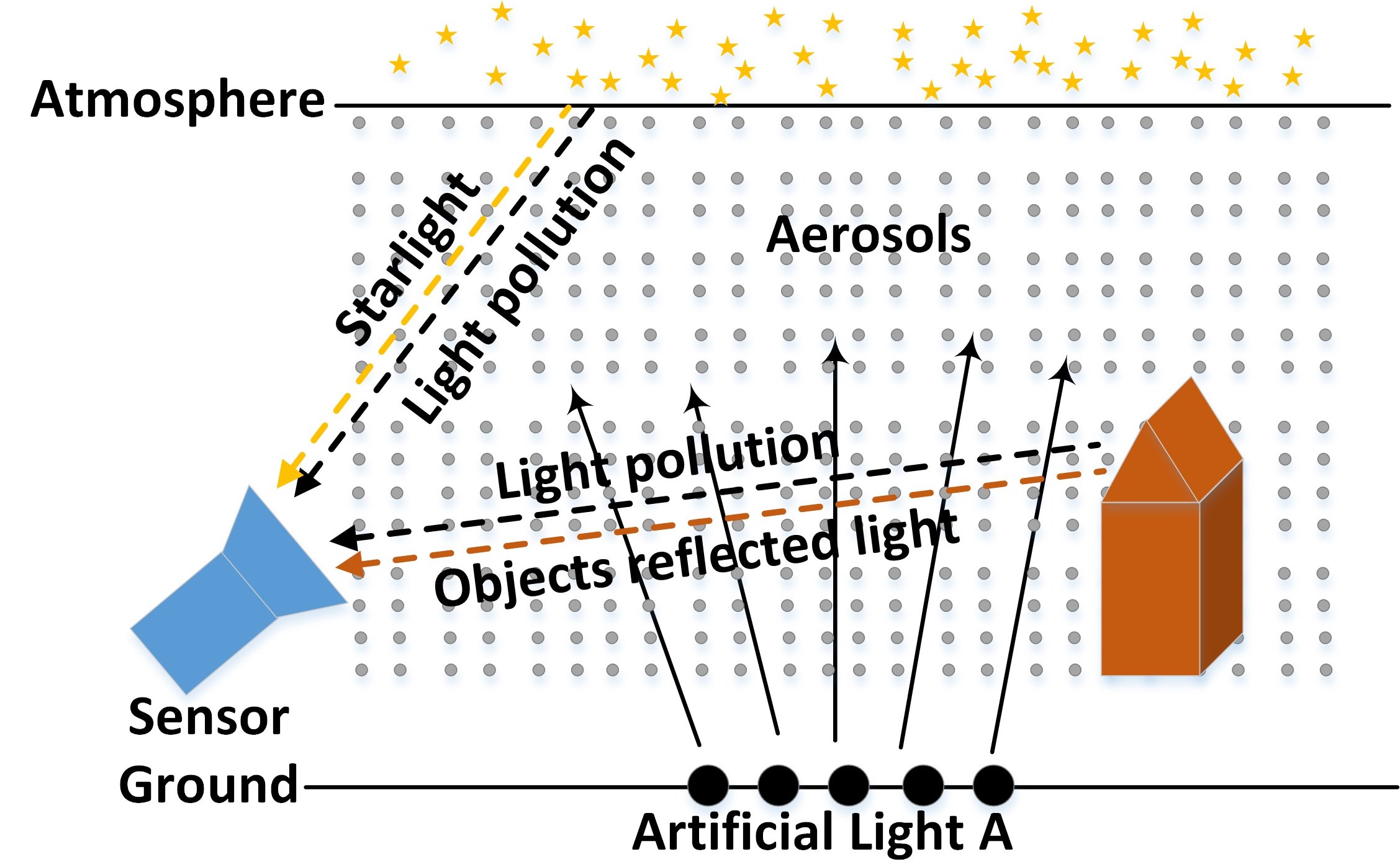}
  \caption{The formation process of light-polluted images.}
\label{fig:Ip_img}
\vspace{-0.5cm}
\end{figure}

The recovery of light pollution free nighttime images is an inverse problem stated below:
\begin{equation}
\hat{I} = I + J
\label{eq:formulation}
\end{equation}
where $\hat{I}$ is the light-polluted image captured by camera, $I$ is the pristine nighttime image that could only be acquired in total void of artificial lights by a perfectly static camera with long exposure, and $J$ is the jamming image formed by artificial lights reflected by aerosols towards the camera. The formation of light-polluted image $\hat{I}$ is schematically depicted in Fig.~\ref{fig:Ip_img}.  Although precise recovery of $I$ or equivalently $J$ from $\hat{I}$ in terms of atmosphere science is very difficult, we aim to develop a practical method that can neutralize light pollution and approximate $I$ in perceptual sense.  To this end, we derive an approximate physical model for the light pollution effect $J$.


The scattering of ground artificial lights by aerosols is the main cause of light pollution. The exact modeling of light pollution is highly complex, if not impossible, as the scattering effects depend on the types, orientations, sizes, and distributions of aerosols permeating the atmosphere, as well as wavelengths, polarization states, and directions of the ground lights \cite{1957Light, mccartney1976optics, narasimhan2002vision, narasimhan2003shedding}. 
We simplify the development of light pollution model by assuming homogeneous atmosphere, namely, aerosols have uniform density and they scatter lights isotropically.

Practical light scattering models seemed to follow the work of  Narasimhan and Nayar \cite{narasimhan2002vision}.
A light gets attenuated as it travels.  Due to aerosol scattering,
a fraction of light flux is removed from the incident beam, and the remaining flux arrived at the destination point is
the attenuated irradiance
given by Bouguer’s exponential law \cite{bouguer1729essai},
\begin{equation}
    E(d,\lambda) = E_0(\lambda) e^{-\beta_\lambda d},
    \label{eq:attenuation}
\end{equation}
where $E_0$ is the radiance of the light source prior to attenuation, $d$ is the distance from the source to destination point, $\lambda$ is the wavelength, and $\beta(\lambda)$ is the scattering coefficient, which accounts for the ability of a unit volume of atmosphere to scatter light of wavelength $\lambda$ in all directions \cite{mccartney1976optics,middleton1957vision}.
For point light sources that radiate isotropically like the street lights with respect to atmosphere, the above attenuation model should be modified to incorporate the inverse-square law,
\begin{equation}
    E(d,\lambda) = \frac{E_0(\lambda) e^{-\beta_\lambda d}}{d^2},
    \label{eq:inverse-square}
\end{equation}

\section{Baseline method}

By light pollution of nighttime images we mean the unwanted effects of ground artificial lights being scattered by aerosols in atmosphere.  To remove visual effects of light pollution, we need to model and compute the light pollution image $J$ so that the pristine image $I = \hat{I} - J$ can be restored.
To simplify the problem, we assume that for each color band $\lambda$, $\lambda \in \{R,G,B\}$, the strength of artificial lighting has a uniform distribution on earth surface, with a constant radiance $A_\lambda$ (a restriction to be removed in the next section).

\begin{figure}[t]
  \centering
  \setlength{\abovecaptionskip}{3pt}
  \includegraphics[width=0.5\linewidth,height=0.35\linewidth]{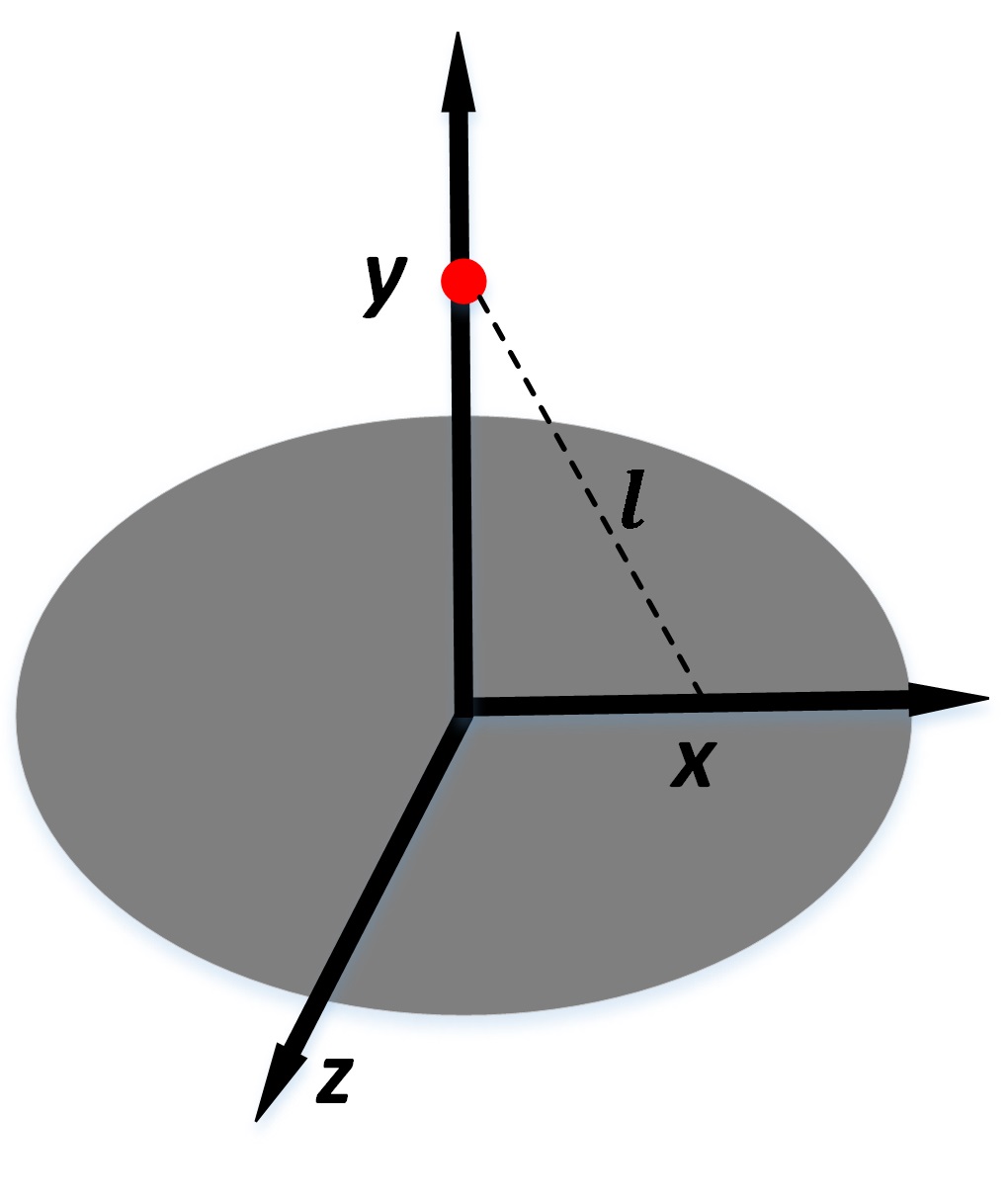}
  \caption{The pollution light irradiance in atmosphere is an integration of the energy that ground artificial lights radiate.}
\label{fig:Ip_h}
\end{figure}

Denote by $E_\lambda(x,y,z)$ the pollution light irradiance of color band $\lambda$ at spatial location $(x,y,z)$.  To keep the image and world coordinates consistent, we let the $y$ axis represent the altitude.
If the pollution lighting has uniform strength and constant color everywhere on ground surface, then $E_{\lambda_0}(x,y_0,z)$ can be considered a constant for any given altitude $y_0$ and wavelength $\lambda_0$.  Therefore, the irradiance function $E_\lambda(x,y,z)$ of artificial lighting is reduced to a univariate function $E_\lambda(y)$ that depends on altitude only, $\lambda \in \{R, G, B\}$.  Using the light attenuation model Eq(\ref{eq:inverse-square}), we compute $E_\lambda(y)$ in the atmosphere by integrating the influxes of ground artificial lights that reach a point of altitude $y$, as illustrated in Fig.~\ref{fig:Ip_h}, and
obtain the radiance of the light pollution at the atmosphere point
\begin{equation}
  E_\lambda(y) = \int_0^\infty \frac{A_\lambda e^{-\beta_\lambda\sqrt{x^2+y^2} }}{x^2+y^2}  2\pi x \, dx.
  \label{eq:Ep(x)}
\end{equation}
With a change of variable $x=\sqrt{l^2-y^2}$, Eq(\ref{eq:Ep(x)}) can be rewritten as
\begin{equation}
  E_\lambda(y) = 2\pi A_\lambda \int_y^\infty \frac{e^{-\beta_\lambda l}}{l} \, dl,
\label{eq:Ep(l)}
\end{equation}
where $l$ is the distance between the ground light source and the atmosphere point of altitude $y$.

\begin{figure}[t]
  \centering
  \captionsetup[subfigure]{belowskip=-0.5\baselineskip}
  \includegraphics[width=0.7\linewidth,height=0.4\linewidth]{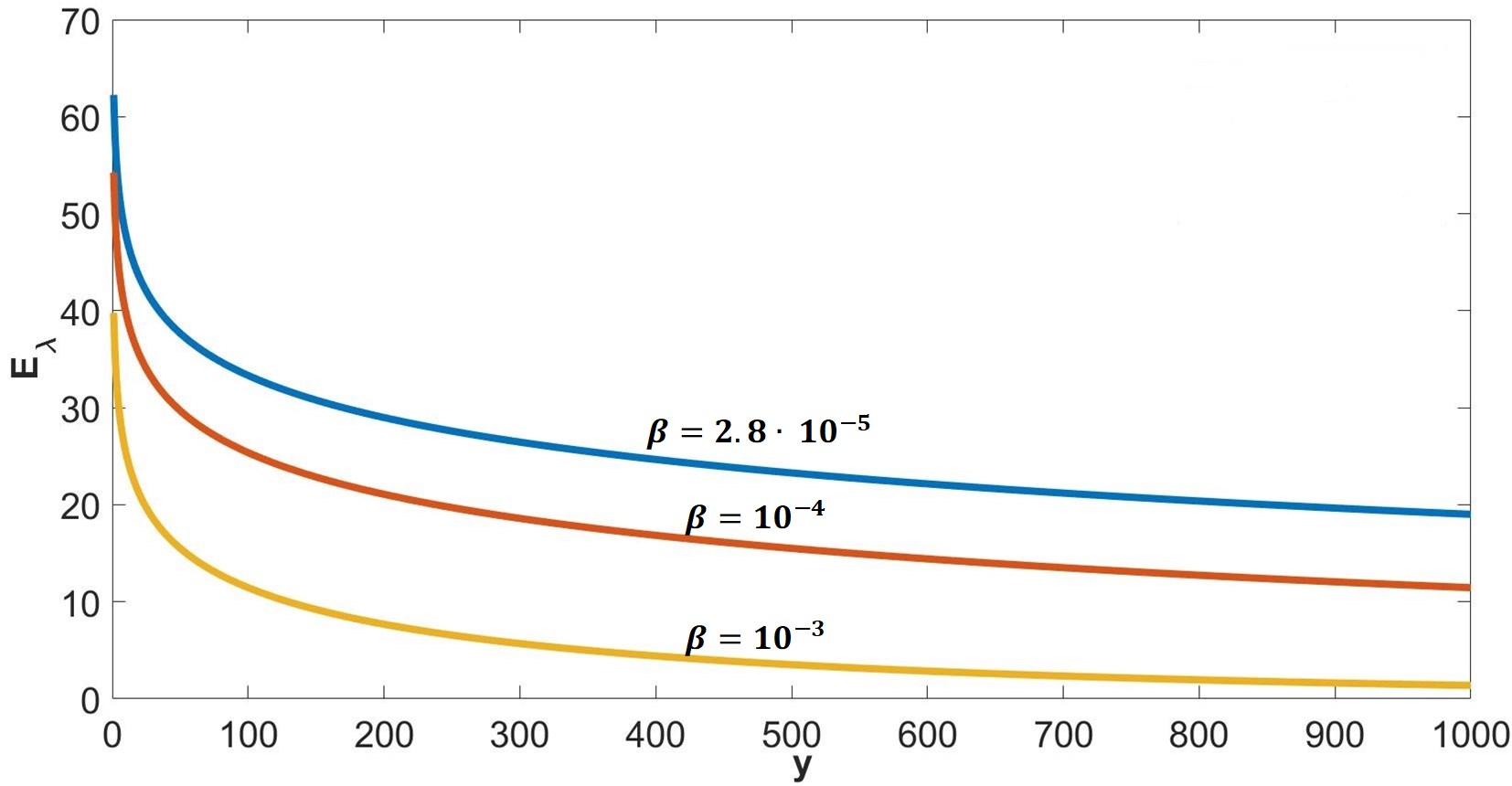}
  \caption{The artificial light irradiance $E_\lambda(y)$ vs.\ the altitude $y$, for different levels of air purity. $2.8 \times 10^{-5}$ represents aerosol free, $10^{-4}$ represents slightly haze, and $10^{-3}$ represents haze.}
\label{fig:Ep_vs_y}
\vspace{-0.4cm}
\end{figure}

Computing Eq(\ref{eq:Ep(l)}) with Taylor expansion yields
\begin{equation}
  E_\lambda(y) = 2\pi A_\lambda \rsx{ \left( \ln l + \sum_{n=1}^\infty \frac{(-\beta_\lambda l)^n}{n \cdot n!} \right) }_{l=y}^\infty
\label{eq:Ep(Taylor)}
\end{equation}
In order to understand how the irradiance of pollution lights varies in altitude and in environment condition, we plot the $E_\lambda(y)$ curves in Fig.~\ref{fig:Ep_vs_y} for different $\beta$ values.  $\beta = 2.8 \cdot 10^{-5}$ corresponds to highly transparent (aerosol free) air, $\beta = 10^{-4}$ to slightly hazy air, and $\beta=10^{-3}$ to haze air \cite{charlson1967direct}.  The curves show that the closer to the ground (the horizon in the image), the higher the level of light pollution.  Such an effect can be observed in light-polluted nighttime images, in which the lower portion of the sky is bathed in scattered ground artificial lights.

\begin{figure}[t]
  \centering
  \captionsetup[subfigure]{belowskip=-0.5\baselineskip}
  \includegraphics[width=0.8\linewidth,height=0.4\linewidth]{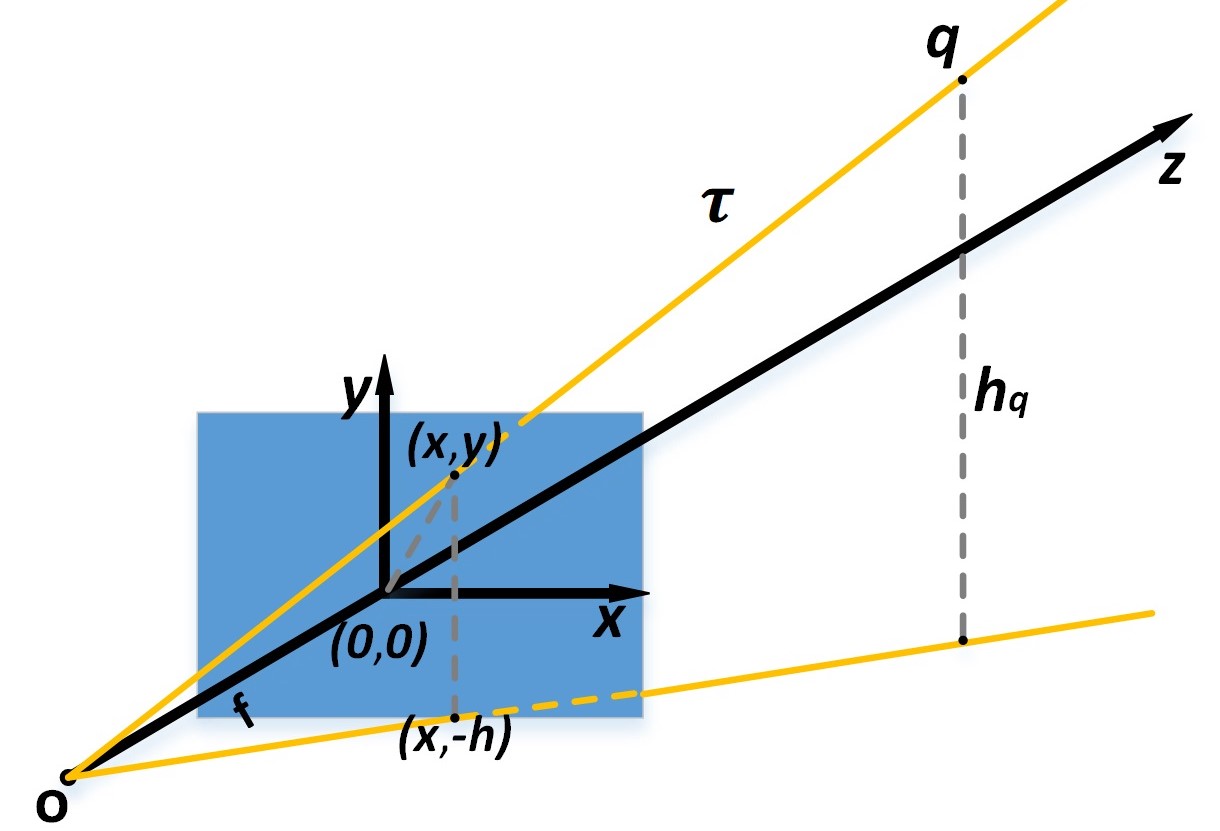}
  \caption{The perspective projection model of a camera. A pixel $(x,y)$ in image $J$ corresponds to a beam of pollution lights towards the camera.}
  \label{fig:Ip_camera}
  \vspace{-0.4cm}
\end{figure}

 Having the energy distribution $E_\lambda(y)$ of unwanted pollution lights in the atmosphere, now we are ready to model the image of light pollution $J$ in Eq(\ref{eq:formulation}).
 A pixel $(x,y)$ in image $J$ corresponds to a beam of pollution lights towards the camera; the pixel value is the accumulation of artificial lights reflected by aerosols along the light pathway.  For point $q$ of distance $\tau$ from the projection center $o$ on the light ray through $(x,y)$ (see Fig.~\ref{fig:Ip_camera}), we have
 \begin{equation}
   h_q = \tau \cdot \frac{y+h}{\sqrt{f^2+x^2+y^2}},
 \end{equation}
 where h is half the height of the image, $f$ is the focal length. Integrating all the artificial lights reflected by aerosols along the light pathway, we obtain the pixel value of pollution image,
 \begin{equation}
   \begin{split}
 J_{\lambda}(x,y) &= \int_0^L E_\lambda \left( \tau \cdot \frac{y+h}{\sqrt{f^2+x^2+y^2}} \right) \beta_\lambda e^{-\beta_\lambda \tau} \, d\tau, \\
 \lambda &\in \{ R,G,B \},
   \end{split}
 \label{eq:Ip_img}
 \end{equation}
 where $L$ is the path length between the sensor pixel $(x,y)$ and the scene point. For pixels in the sky, $L$ is set to infinity.  Once having $J_\lambda$ computed, the baseline LPR algorithm estimates the pollution-free image $I_\lambda$ to be $\hat{I}_\lambda - J_\lambda$.

\section{Adaptive method}

In the previous section, we oversimplified the source of light pollution to be an artificial light emitting surface of uniform strength and constant color.
Unlike the uniform sun light that is scattered by aerosols to cause haze, artificial lights in the night have, in general, an uneven geographical distribution.  Therefore, in each color band $\lambda \in \{R,G,B\}$, the radiance of artificial lights is a spatially varying function $A_{\lambda}(x,z)$, not a constant $A_\lambda$ as in Eq(\ref{eq:Ep(x)}).
Accordingly, for better LPR results we need to improve the above baseline algorithm by making it spatially adaptive.

It is very difficult to compute the 2D radiance function $A_{\lambda}(x,z)$ from input image $\hat{I}$, because $\hat{I}$ offers very little information in the $z$ direction.  The next best and feasible step is to model the horizontal variations of $A_{\lambda}(x,z)$, or the marginal distribution of pollution light radiance along the horizon.  Projecting $A_{\lambda}(x,z)$ to the $x$ axis and reducing it to $A_{\lambda}(x)$ is acceptable, because the most common composition of nighttime photos is a horizontal landscape and thus the $x$ axis is the principal axis of the 2D function $A_{\lambda}(x,z)$.

After the above simplification, the ground artificial lights can be seen as collimated light sources.  The upward pollution light rays have irradiance decay according to Eq(\ref{eq:attenuation}).  At altitude $y$ the artificial light radiance $A_\lambda(x)$ is reduced to
\begin{equation}
E_\lambda(x,y) = A_\lambda(x)e^{-\beta_\lambda y}.
\label{Ep(x,y)}
\end{equation}
Substituting $E_\lambda(x,y)$ in Eq(\ref{eq:Ip_img}), we obtain the pollution image,
\begin{equation}
  \setlength\belowdisplayskip{0.8pt}
  \begin{split}
  J_{\lambda}(x,y) &= A_\lambda(x) \cdot \alpha(x,y), \cr
  \alpha(x,y) &= \int_0^L e^{-\beta_\lambda \tau \cdot \frac{y+h}{\sqrt{f^2+x^2+y^2}}} \beta_\lambda e^{-\beta_\lambda \tau} \, d\tau. \cr
  \label{eq:J(x,y)_ada}
  \end{split}
\end{equation}

\begin{figure}[t]
  \centering
    \begin{subfigure}{0.46\linewidth}
      \centering
      \includegraphics[width=\linewidth,height=0.66\linewidth]{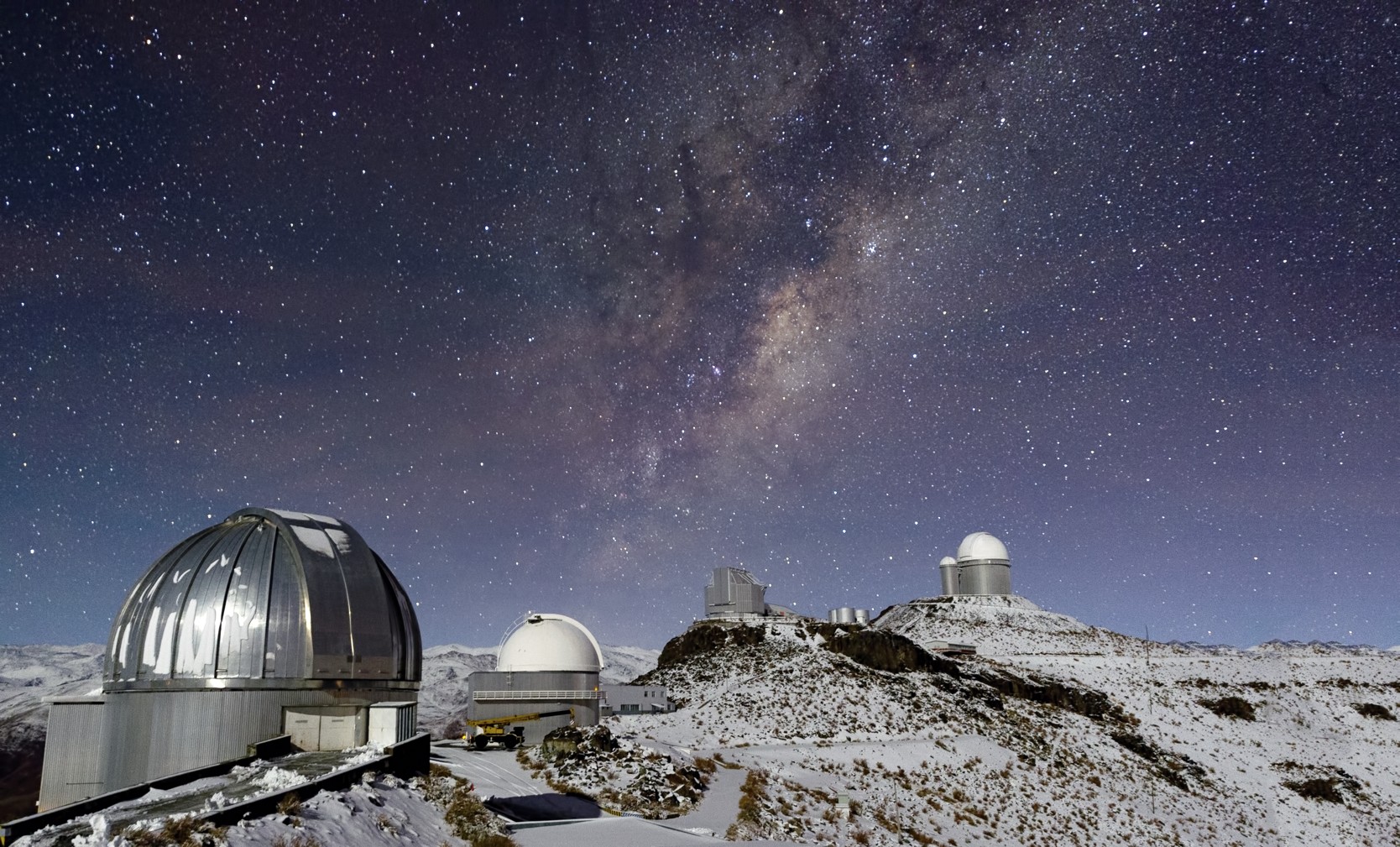}
    \end{subfigure}
    \begin{subfigure}{0.46\linewidth}
      \centering
      \includegraphics[width=\linewidth,height=0.66\linewidth]{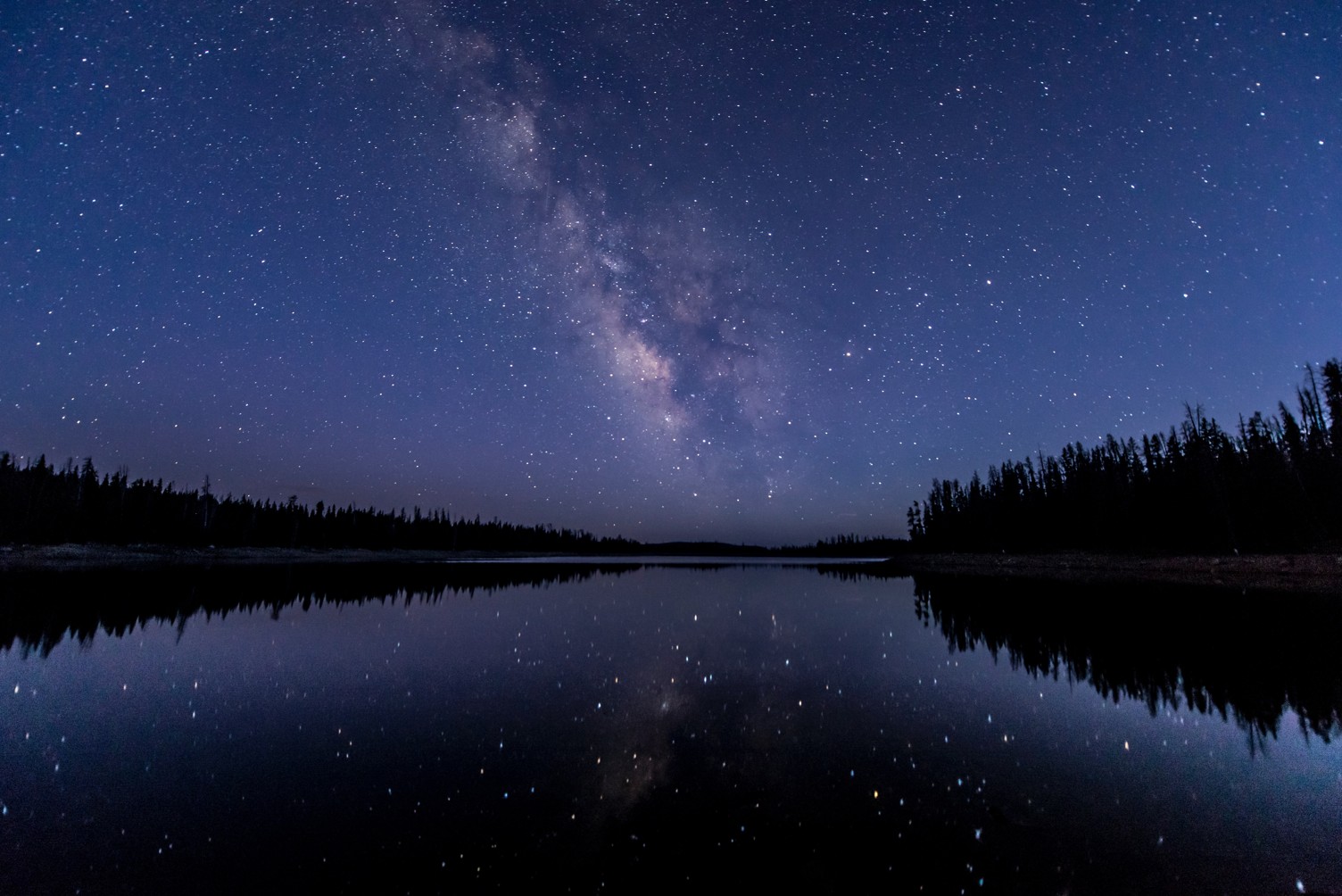}
    \end{subfigure}

    \begin{subfigure}{0.46\linewidth}
      \centering
      \includegraphics[width=\linewidth,height=0.66\linewidth]{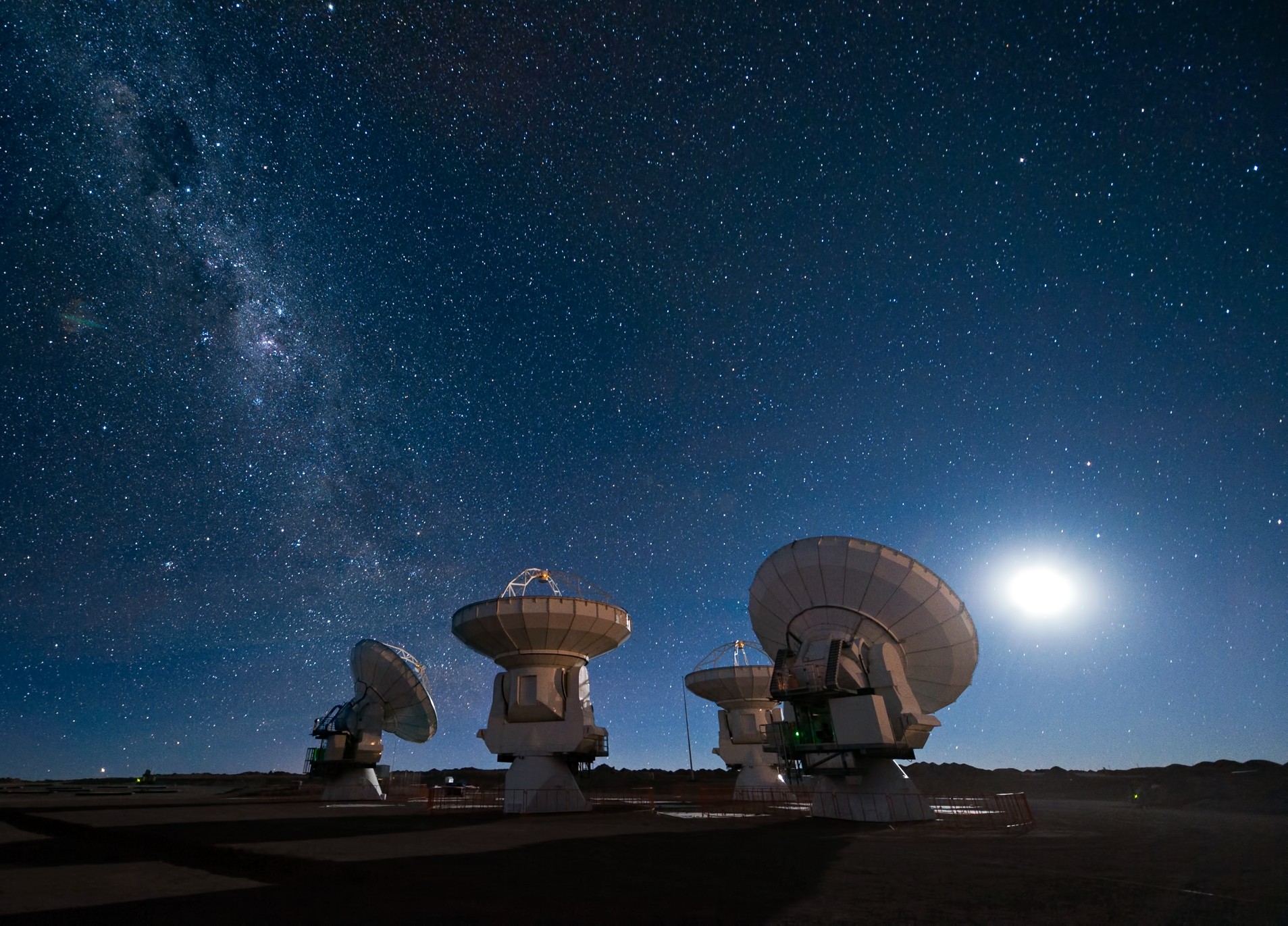}
    \end{subfigure}
    \begin{subfigure}{0.46\linewidth}
      \centering
      \includegraphics[width=\linewidth,height=0.66\linewidth]{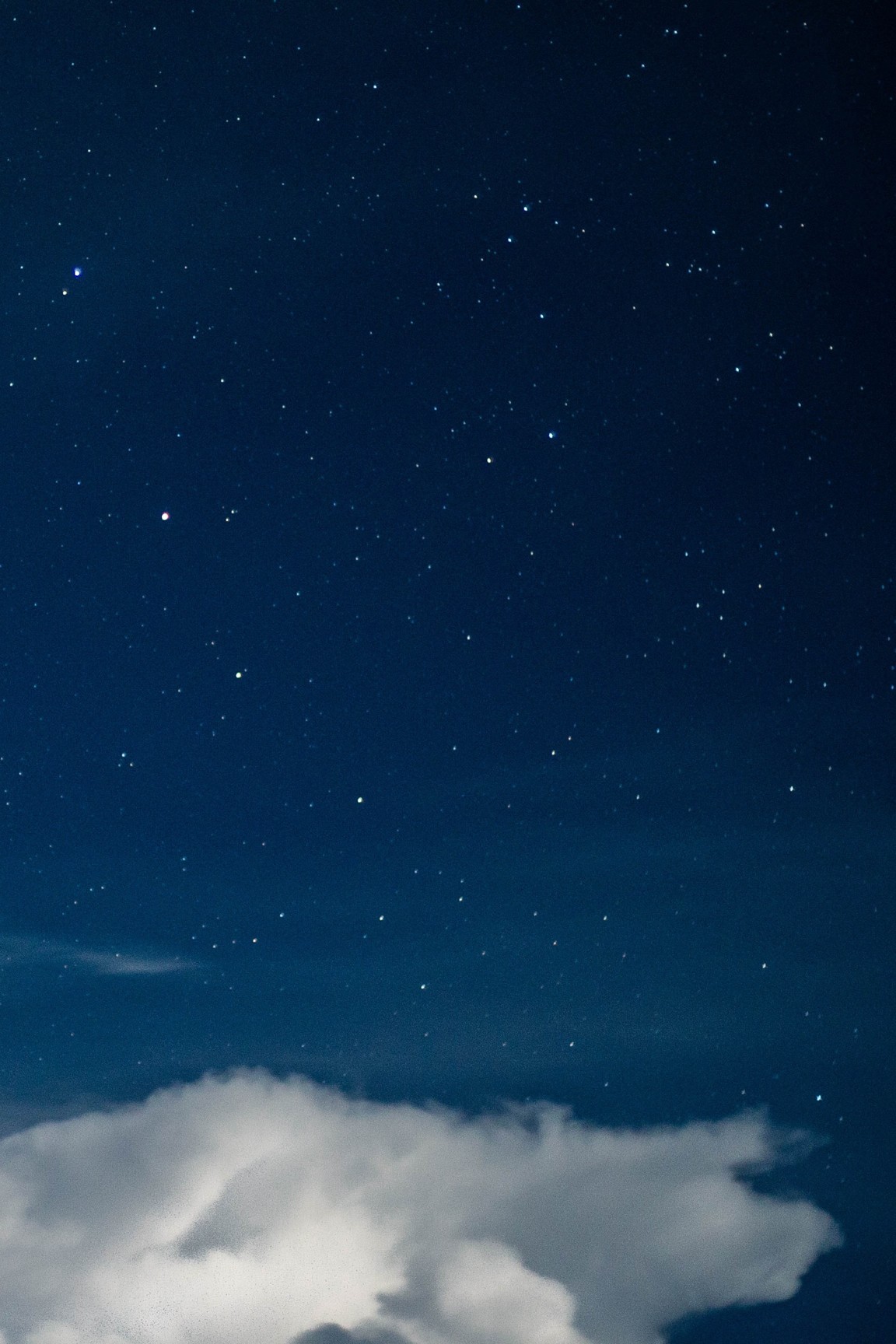}
    \end{subfigure}
  \caption{Long-exposure nighttime sky images captured in environments free of artificial lights \cite{night_sky_clean2,night_sky_clean1}.}
\label{fig:clean_sky}
\vspace{-0.4cm}
\end{figure}

To compute the pollution image $J_\lambda(x,y)$ using Eq(\ref{eq:J(x,y)_ada}), we need to know the horizontal radiance profile $A_{\lambda}(x)$ of ground pollution lights, and separately in color bands $\lambda \in \{R,G,B\}$, i.e., know the spectral signature of the pollution lights.  Now we develop a method to estimate $A_{\lambda}(x)$ by starting from some known priors on a pure night sky without artificial lights and working its way backward.
If there was a total absence of artificial lights, then the latent image $I_{\lambda}(x,y)$ of the night sky would have low intensity; more importantly, a horizontal strip $I_{\lambda}(x,y_0)$ far above ground should be almost a constant, given the altitude $y_0$ and the color band $\lambda$.  Therefore, if the horizontal strip $\hat{I}_{\lambda}(x,y_0)$ of the input image is nonuniform, then $\hat{I}_{\lambda}(x,y_0)$ reflects the spatial radiance distribution of ground artificial lighting. This gives us a clue to estimate the required spatial distribution $A_{\lambda}(x)$.

It follows from Eq(\ref{eq:J(x,y)_ada}) and $J_\lambda=\hat{I}_\lambda-I$ that
\begin{equation}
  A_{\lambda}(x) = \frac{\hat{I}_{\lambda}(x,y)-I_{\lambda}(x,y)}{\alpha(x,y)},
  \lambda \in \{ R,G,B \}.
\label{eq:artificial_lighting}
\end{equation}
In other words, $A_{\lambda}(x)$ can be derived from the input image $\hat{I}_{\lambda}(x,y)$ as long as if the latent image $I_{\lambda}(x,y)$ is known for some altitude $y=y_0$ sufficiently high above.  The required priors are not difficult to be drawn from the relatively large number of night sky images free of artificial light pollution that are available from various sources, including the Internet.  Samples of such pristine night sky images are presented in Fig.~\ref{fig:clean_sky}.

\begin{figure}[t]
  \centering
  \setlength{\abovecaptionskip}{4pt}
    \begin{subfigure}{0.48\linewidth}
      \centering
      \includegraphics[width=\linewidth,height=0.5\linewidth]{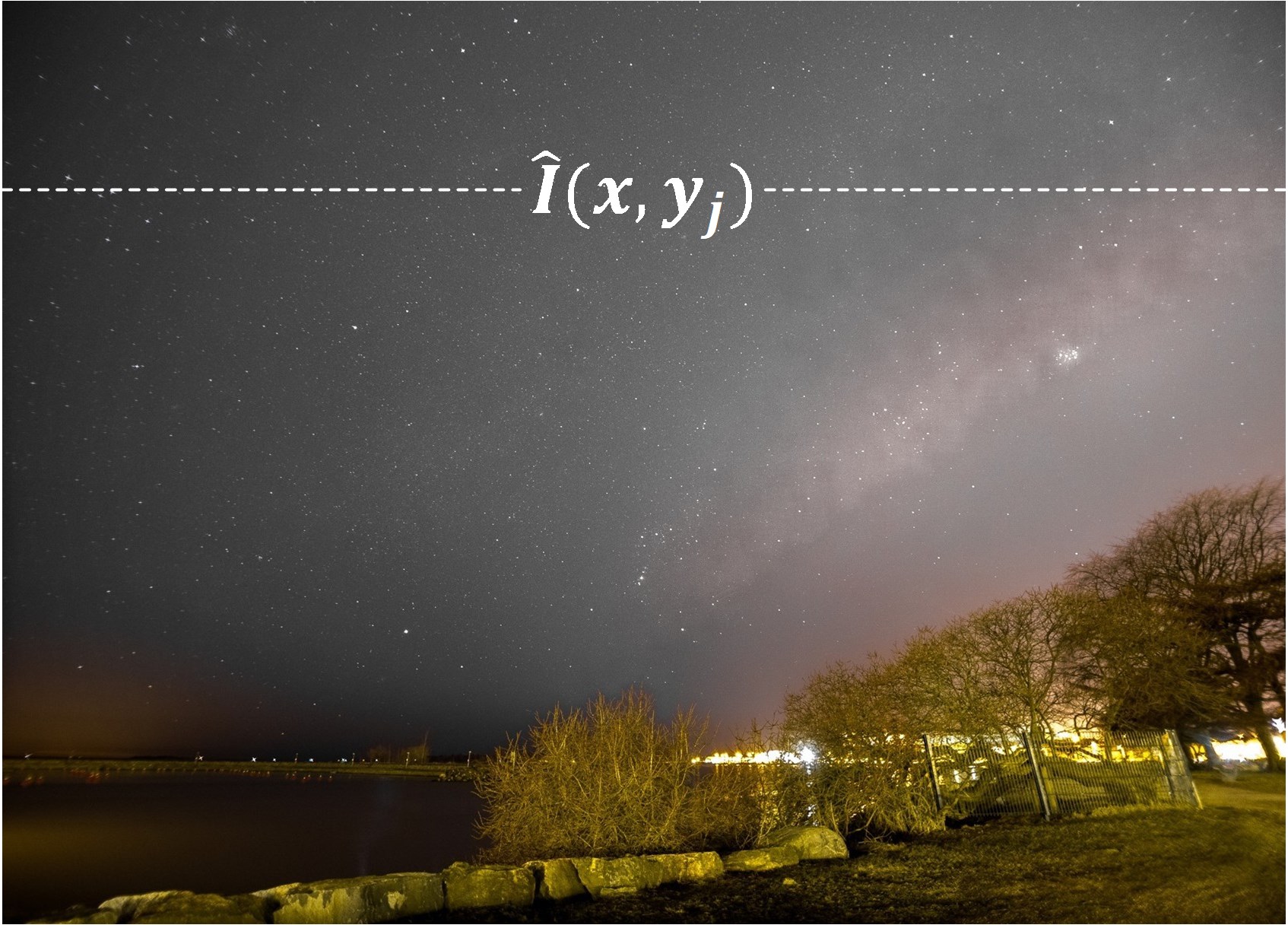}
      \caption{}
    \end{subfigure}
    \begin{subfigure}{0.48\linewidth}
      \centering
      \includegraphics[width=\linewidth,height=0.5\linewidth]{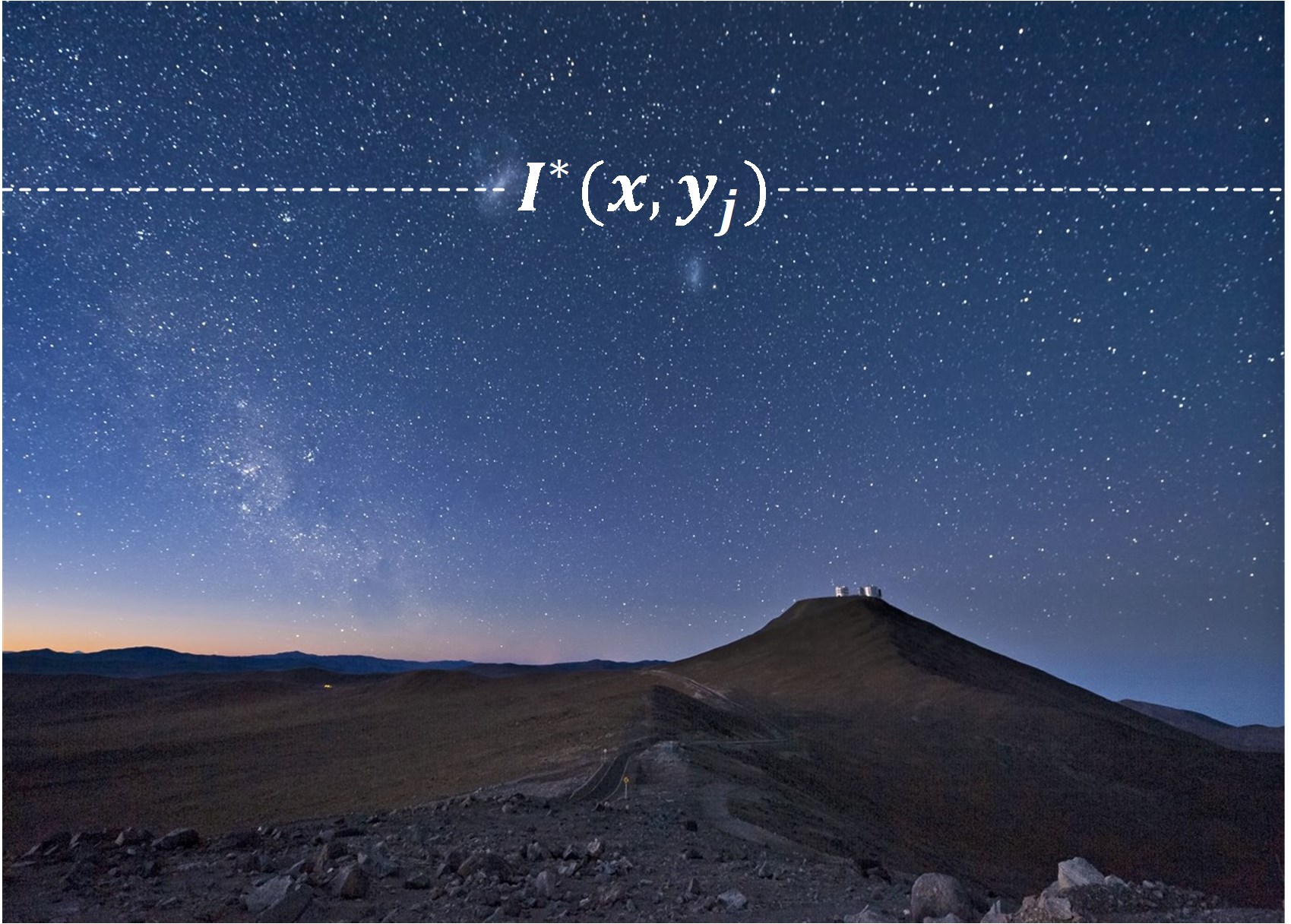}
      \caption{}
    \end{subfigure}

    \begin{subfigure}{0.48\linewidth}
      \centering
      \includegraphics[width=\linewidth,height=0.5\linewidth]{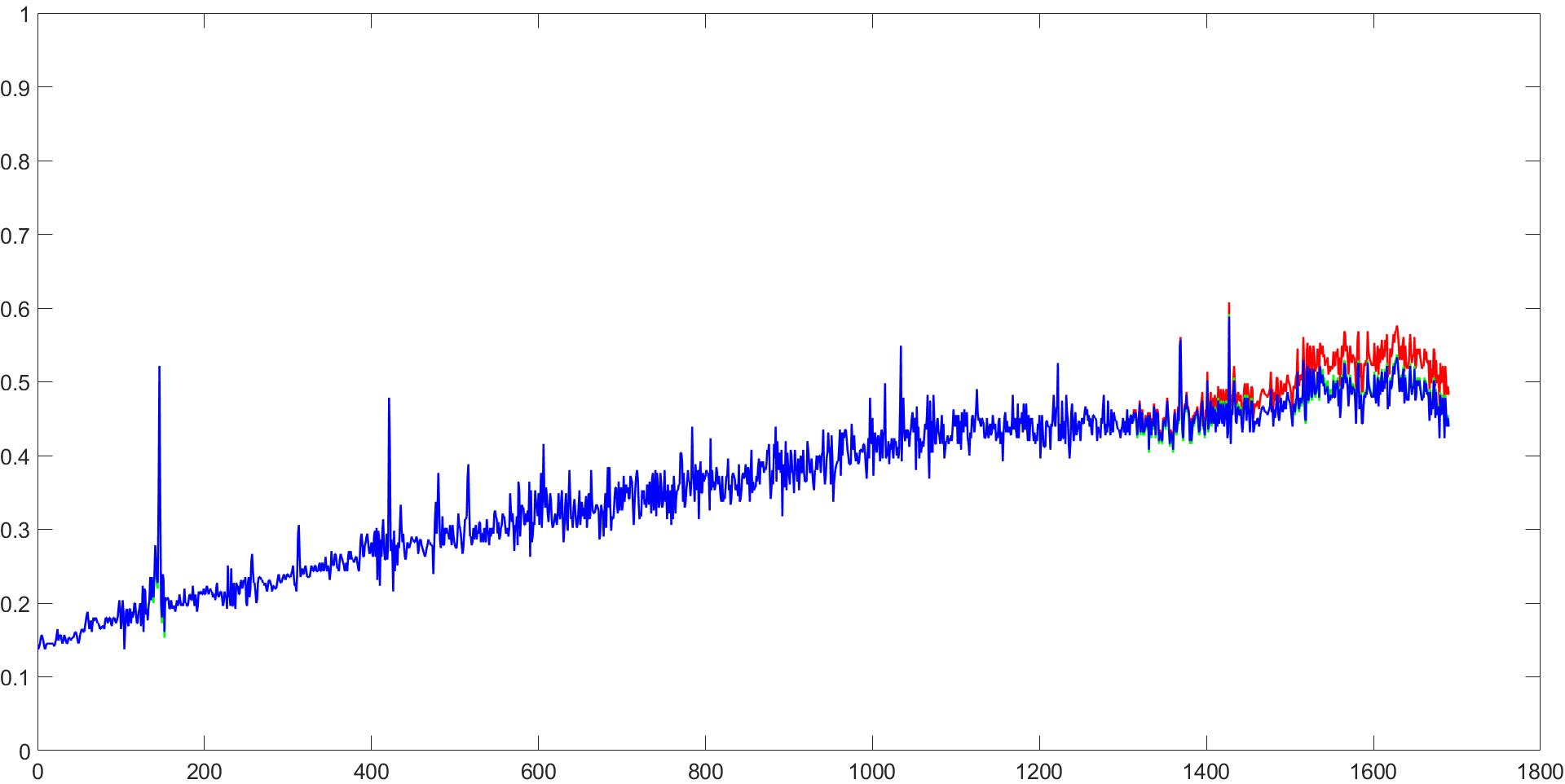}
      \caption{}
    \end{subfigure}
    \begin{subfigure}{0.48\linewidth}
      \centering
      \includegraphics[width=\linewidth,height=0.5\linewidth]{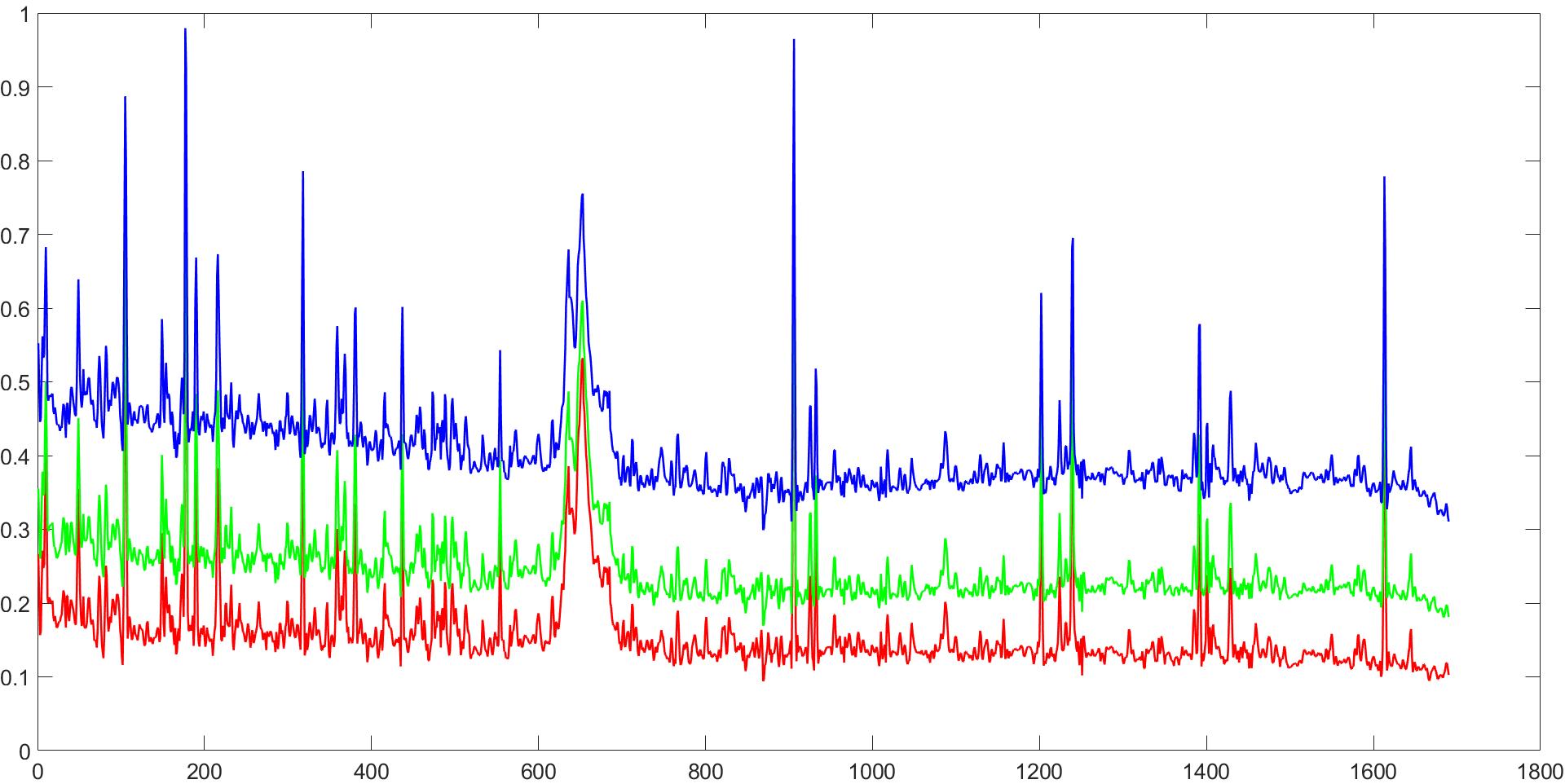}
      \caption{}
    \end{subfigure}

    \begin{subfigure}{0.48\linewidth}
      \centering
      \includegraphics[width=\linewidth,height=0.5\linewidth]{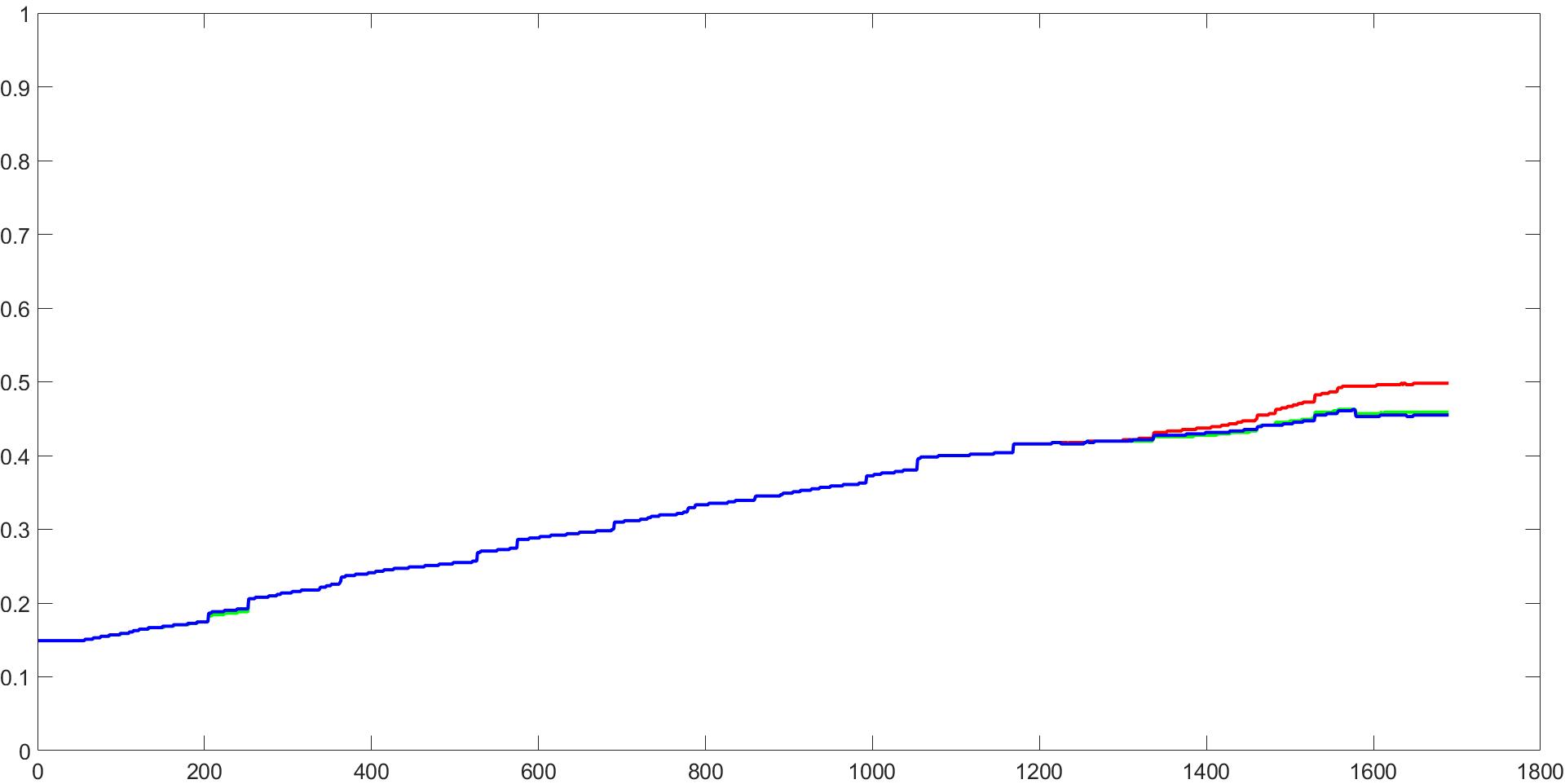}
      \caption{}
    \end{subfigure}
    \begin{subfigure}{0.48\linewidth}
      \centering
      \includegraphics[width=\linewidth,height=0.5\linewidth]{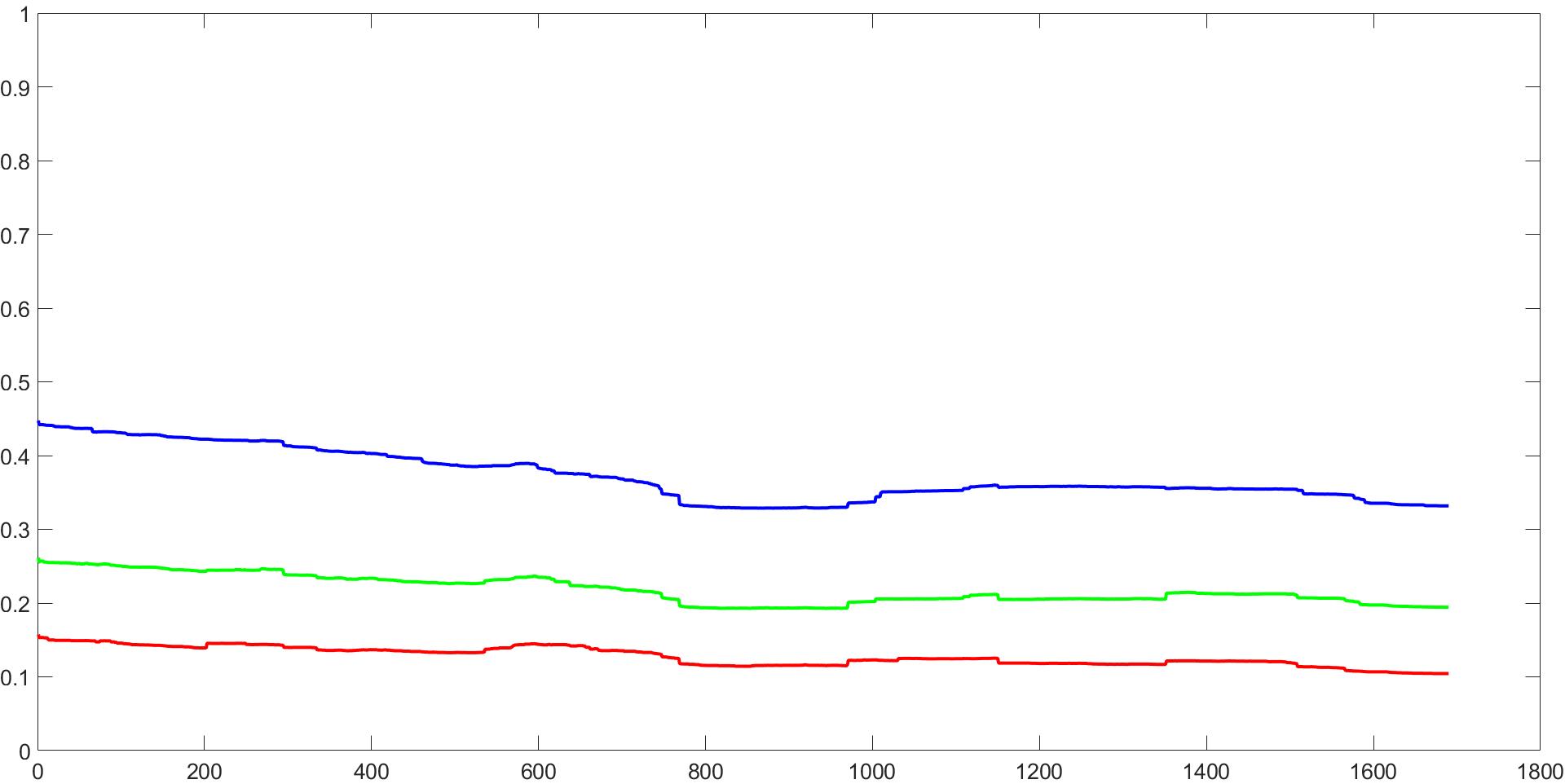}
      \caption{}
    \end{subfigure}
  \caption{(a): Light-polluted image with spatially nonuniform radiance distribution of artificial lights. (b): Nighttime light pollution-free image for calibration. (c) and (d): 1D signals (in R, G, B) of pixel rows marked in (a) and (b). (e) and (f): The quasi-quartile filtered results of (c) and (d).}
\label{fig:horizontal_strip}
\vspace{-0.4cm}
\end{figure}

In nature even without artificial lighting, the night sky is still illuminated by following natural compounded sources: the Moon that reflects the sunlight; the Sun that is set but its light is still scattered around the edge of the Earth (a.k.a., astronomical twilight); The planets and stars; the zodiacal light; airglow.  The last three account for significant portions of illumination in the moonless night sky.  But only by sufficiently long exposures, the effects of the above weak natural light sources can be clearly imaged as we see in Fig.~\ref{fig:clean_sky}.  Indeed,
some researchers found that via long exposure the imaged night sky in truly dark environment appears blue like in the daytime \cite{night_sky_clean1,shaw1996color}.
We use these images to calibrate our restoration method precisely because we want to reproduce the visual appeal of long-exposure night photography in urban surroundings without the side effects of artificial light pollution.

To proceed with the above idea, we need to roughly align the sky portions of the input image $\hat{I}_{\lambda}(x,y)$ and a chosen light pollution-free latent image $I^*_{\lambda}(x,y)$, called calibration image.  This can be accomplished by using one of many skyline detection algorithms \cite{frajberg2017convolutional,hung2013skyline,lie2005robust} and scaling, if needed.
To make the estimation of $A_{\lambda}(x)$ more robust,
we choose a set $\mathbb{Y}$ of several pixel rows in the sky far above horizon.
The input image and calibration image are cross examined at these pixel rows $y_j$, $j \in \mathbb{Y}$, to estimate the spatial distribution of pollution radiance $A(x)$.

In order to prevent the lights of stars and Moon from interfering the estimation of $A_{\lambda}(x)$, we filter both 1D signals $\hat{I}(x,y_j)$ and $I^*(x,y_j)$ using a so-called quasi-quartile filter
\begin{equation}
  \begin{split}
\hat{I}_{1/4}(x,y_j) &= [min(\hat{I}(x,y_j))+median(\hat{I}(x,y_j))]/2, \cr
I^*_{1/4}(x,y_j) &= [min(I^*(x,y_j))+median(I^*(x,y_j))]/2,
  \end{split}
\end{equation}
where $min$ and $median$ are the minimum and median filters in the $x$ direction.  The filtered results $\hat{I}_{1/4}(x,y_j)$ and $I^*_{1/4}(x,y_j)$ are sky background pixel rows in presence and absence of artificial lights.  The selection of calibration pixel rows and the role of the quasi-quartile filter are depicted in Fig.~\ref{fig:horizontal_strip}.  Note how the night sky spectral signatures differ between the light-polluted and pure calibration images.
Now we are ready to solve the following optimization problem to obtain $A_{\lambda}(x)$,
\begin{equation}
A(x) = \arg \min_{z} \sum_{j\in \mathbb{Y}}
\left[ z - \frac{\hat{I}_{1/4}(x,y_j) - I^*_{1/4}(x,y_j)}{\alpha(x,y_j)} \right]^2.
\end{equation}

\section{Restoration of light-polluted city images}

\begin{figure}[t]
   \centering
   \setlength{\abovecaptionskip}{4pt}
     \begin{subfigure}{0.45\linewidth}
       \centering
       \includegraphics[width=\linewidth,height=0.66\linewidth]{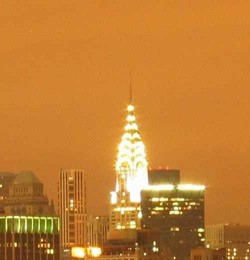}
       \caption{Light-polluted image}
     \end{subfigure}

     \begin{subfigure}{0.45\linewidth}
       \centering
       \includegraphics[width=\linewidth,height=0.66\linewidth]{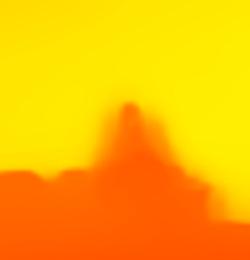}
       \caption{Depth map by \cite{xian2020structure}}
     \end{subfigure}
     \begin{subfigure}{0.45\linewidth}
       \centering
       \includegraphics[width=\linewidth,height=0.66\linewidth]{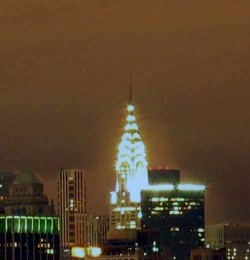}
       \caption{Restoration result with (b) }
     \end{subfigure}

     \begin{subfigure}{0.45\linewidth}
       \centering
       \includegraphics[width=\linewidth,height=0.66\linewidth]{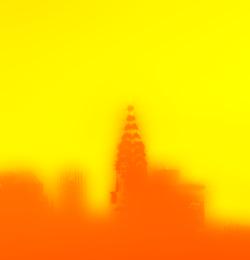}
       \caption{Guided filtered depth map}
     \end{subfigure}
     \begin{subfigure}{0.45\linewidth}
       \centering
       \includegraphics[width=\linewidth,height=0.66\linewidth]{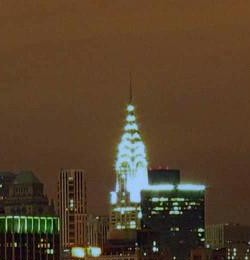}
       \caption{Restoration result with (d)}
     \end{subfigure}

   \caption{Restoration results of a light-polluted city image (a) with an estimated depth map (b) and the edge-guided filtered depth map (d).  Note the removal of halos around the skyline from image (c) to image (e).}
 \label{fig:halos}
 \vspace{-0.4cm}
 \end{figure}

 \begin{figure*}[t]
   \centering
     \begin{subfigure}{0.32\linewidth}
       \centering
       \includegraphics[width=\linewidth,height=0.48\linewidth]{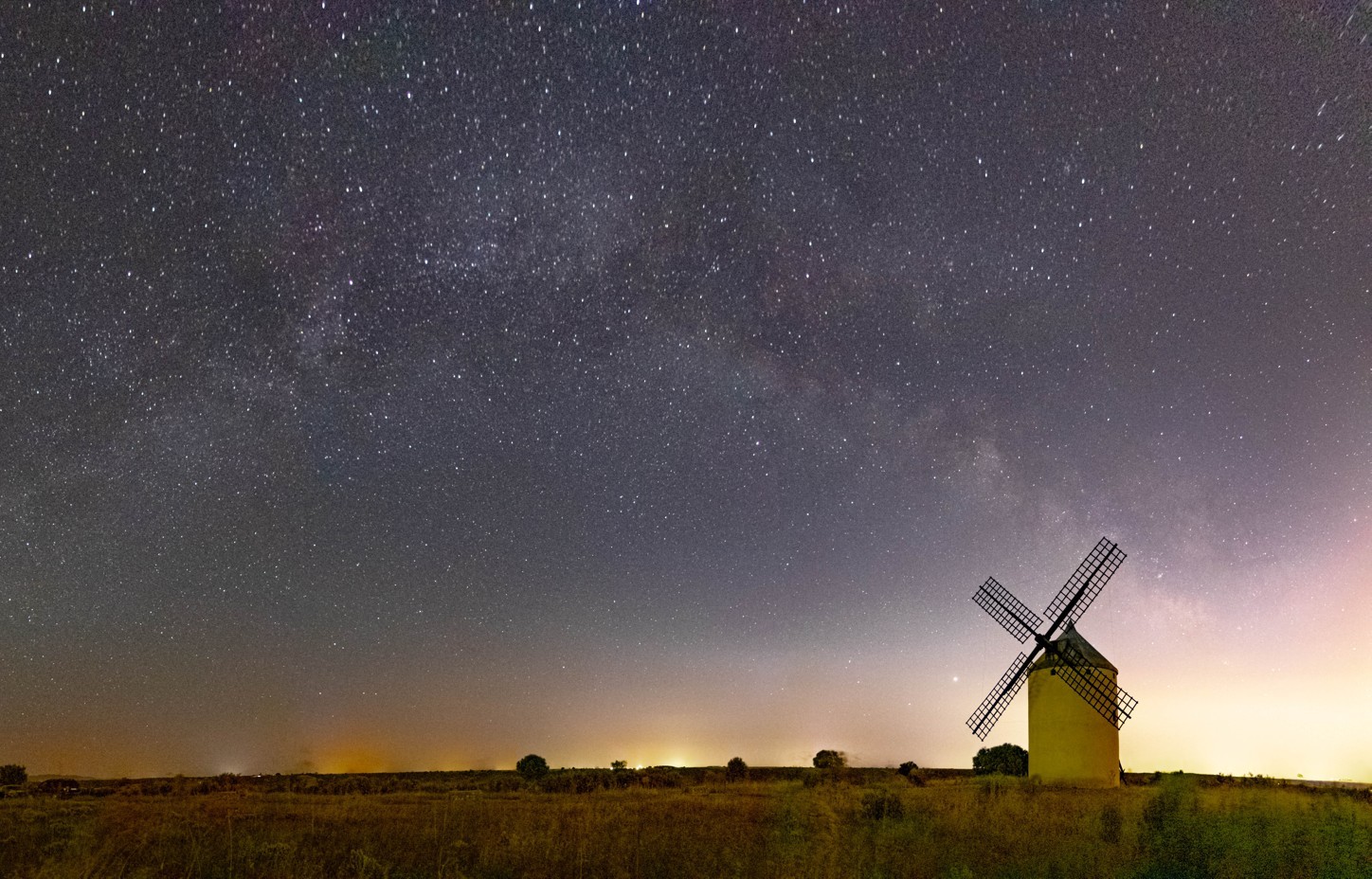}
       \caption{Light-polluted image}
     \end{subfigure}
     \begin{subfigure}{0.32\linewidth}
       \centering
       \includegraphics[width=\linewidth,height=0.48\linewidth]{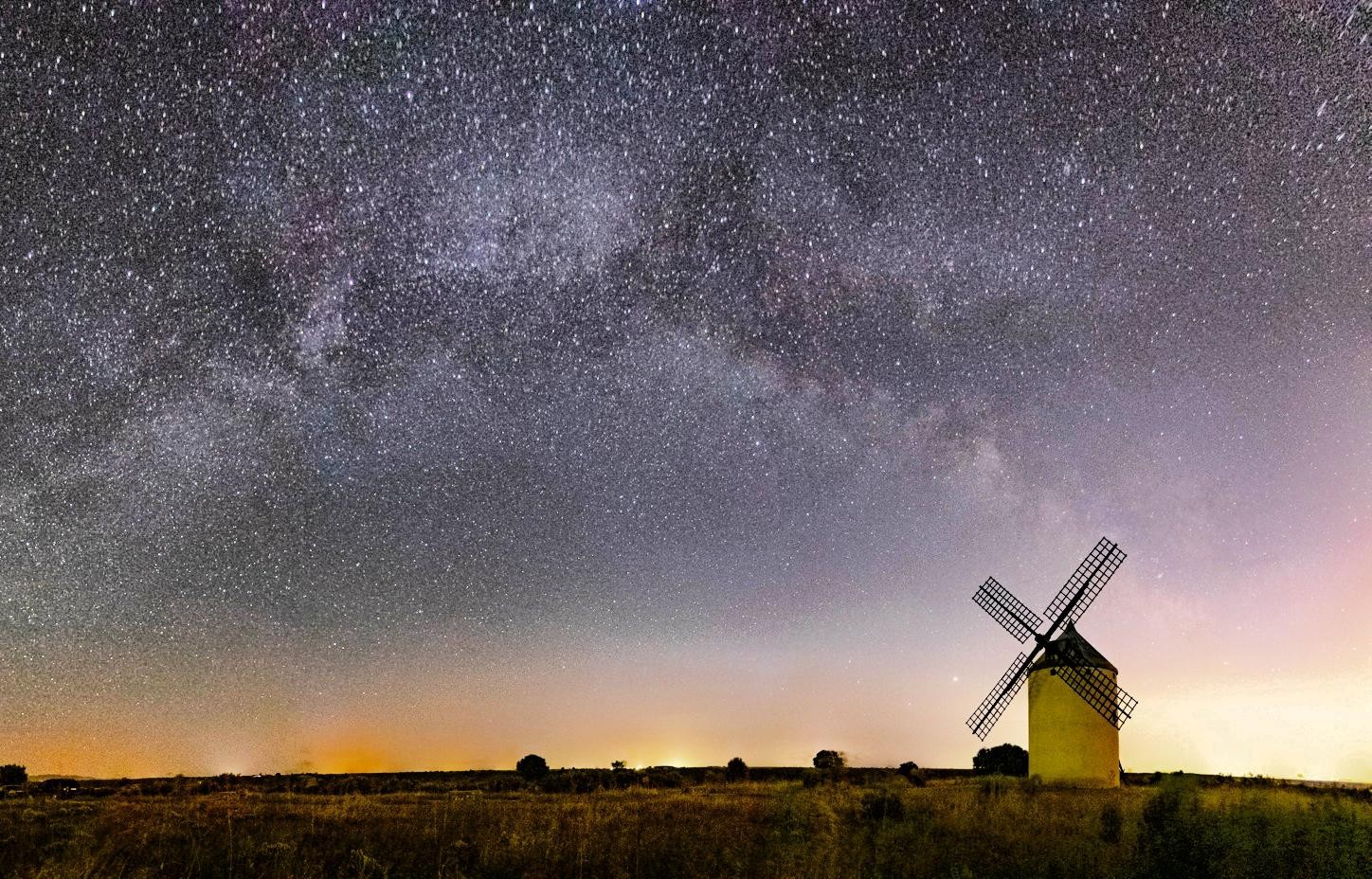}
       \caption{CLAHE}
     \end{subfigure}
     \begin{subfigure}{0.32\linewidth}
       \centering
       \includegraphics[width=\linewidth,height=0.48\linewidth]{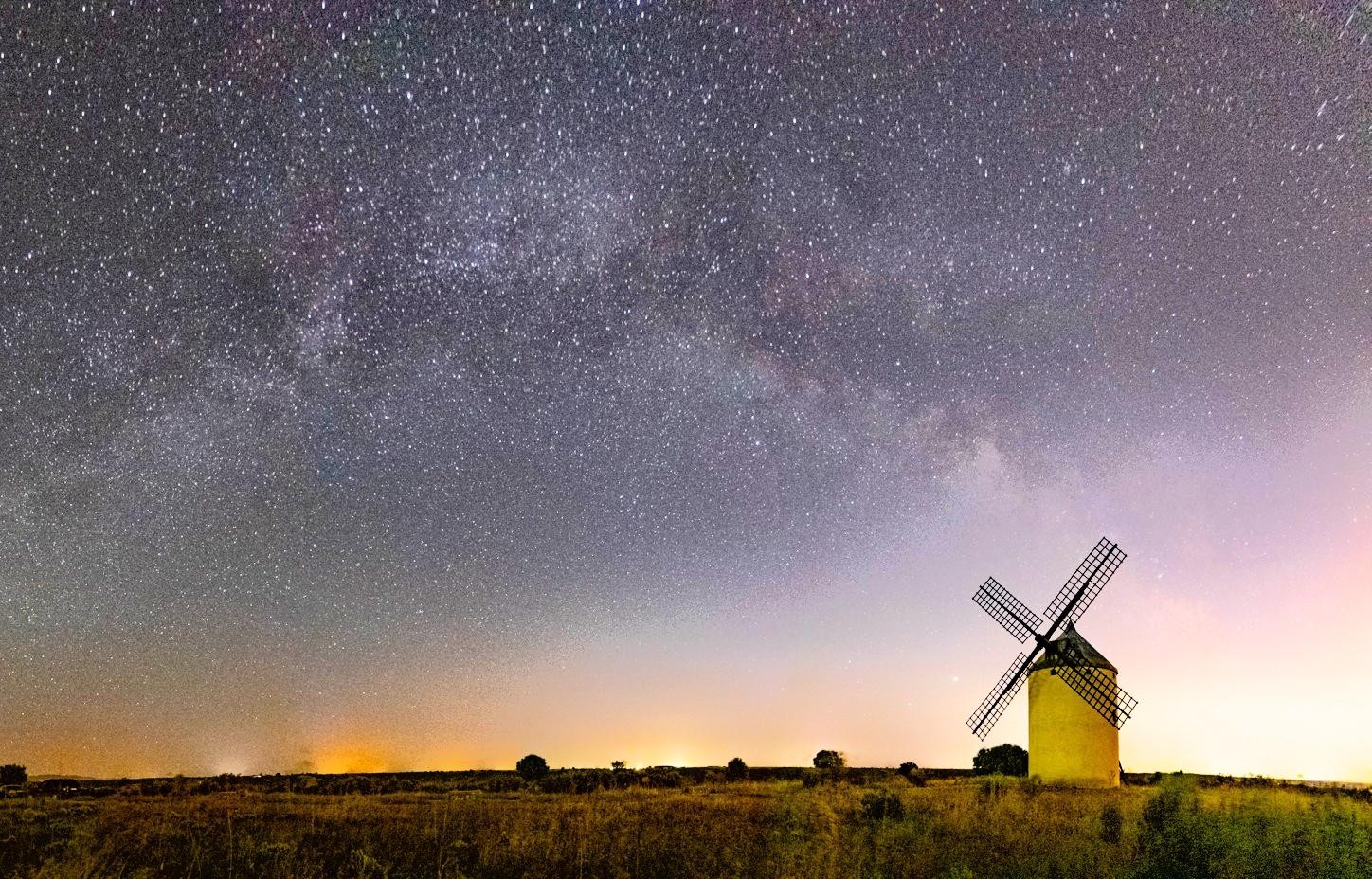}
       \caption{OCTM}
     \end{subfigure}

     \begin{subfigure}{0.32\linewidth}
       \centering
       \includegraphics[width=\linewidth,height=0.48\linewidth]{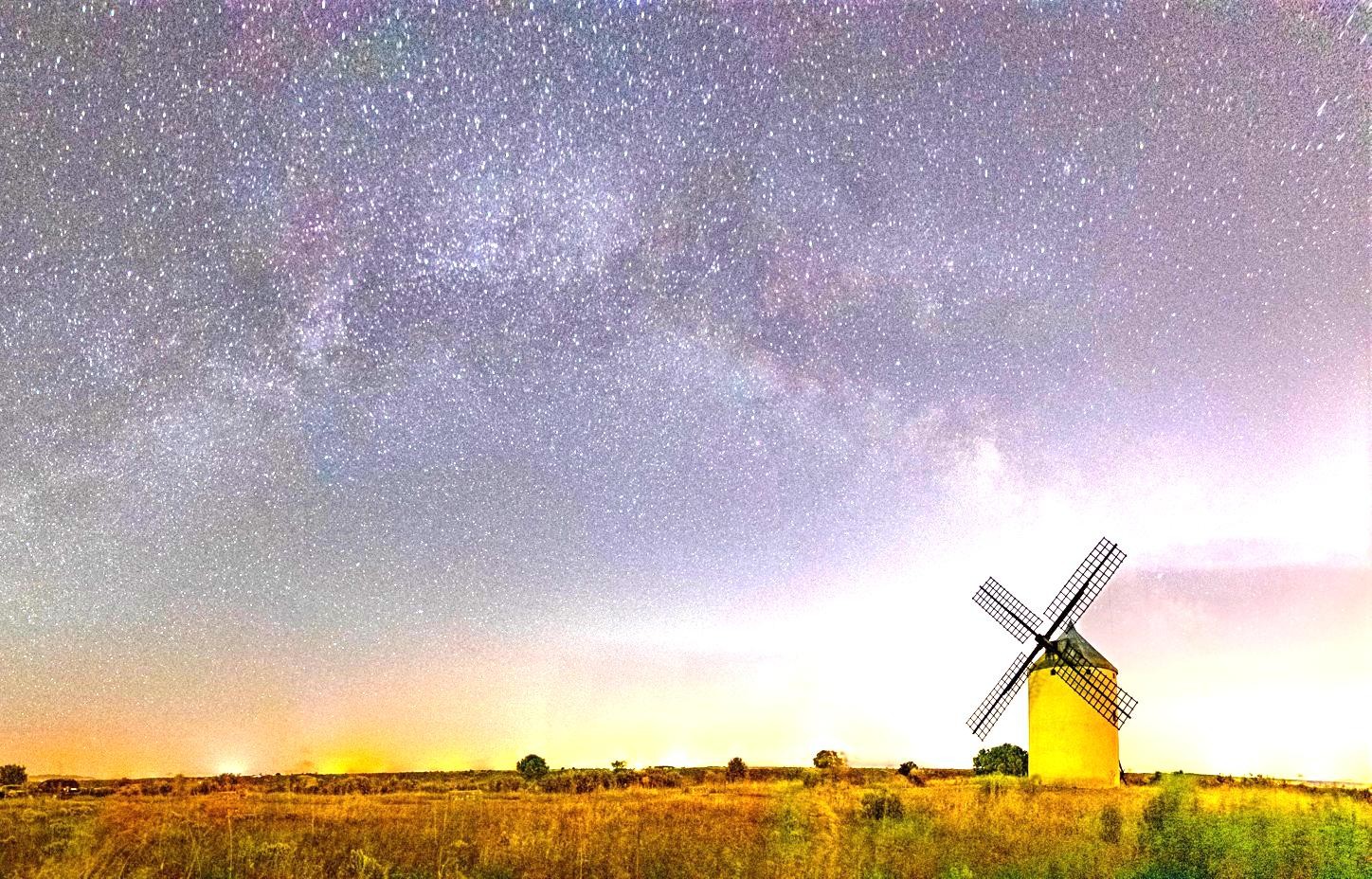}
       \caption{LIME}
     \end{subfigure}
     \begin{subfigure}{0.32\linewidth}
       \centering
       \includegraphics[width=\linewidth,height=0.48\linewidth]{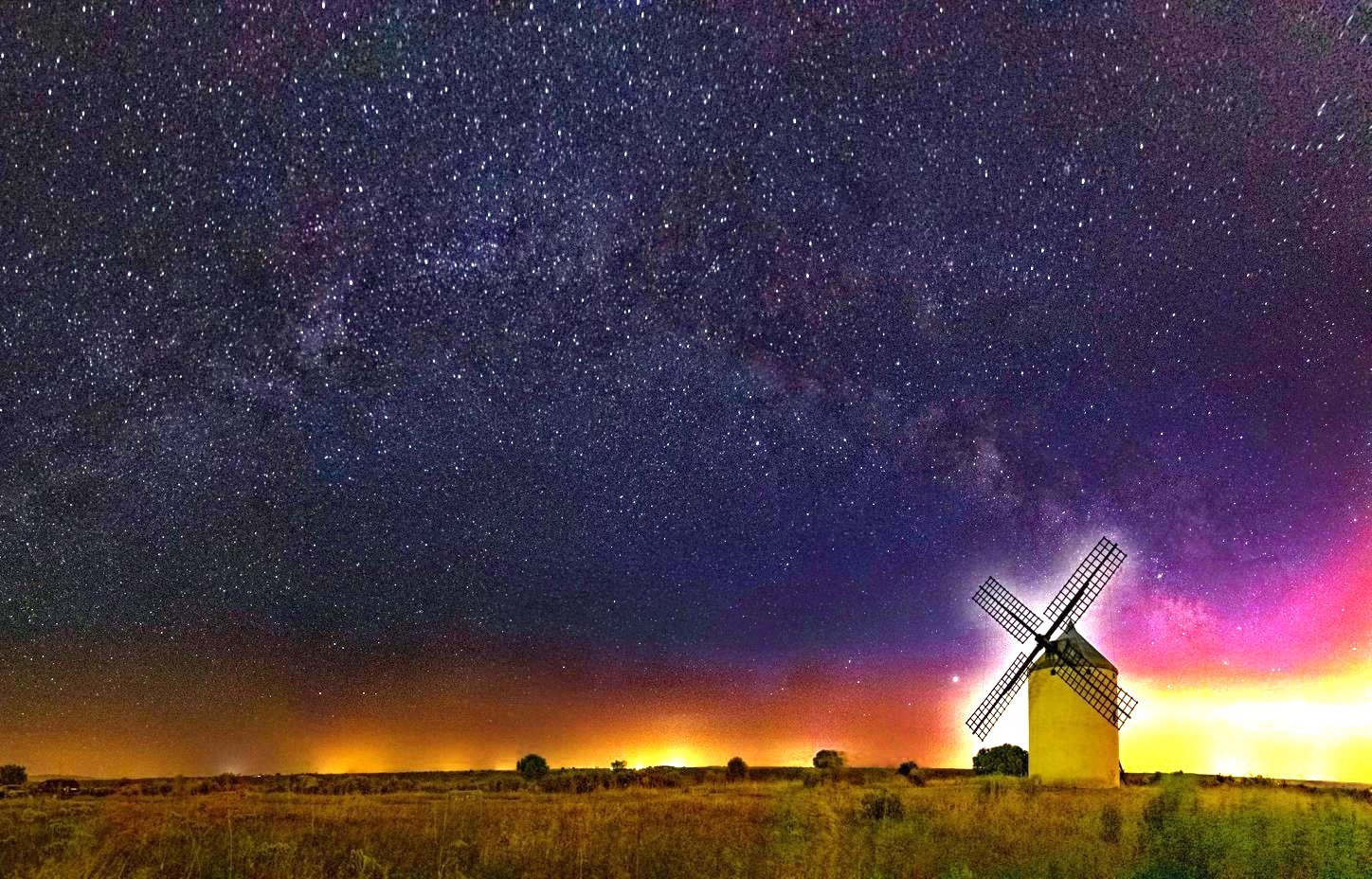}
       \caption{Dark channel dehazing}
     \end{subfigure}
     \begin{subfigure}{0.32\linewidth}
      \centering
      \includegraphics[width=\linewidth,height=0.48\linewidth]{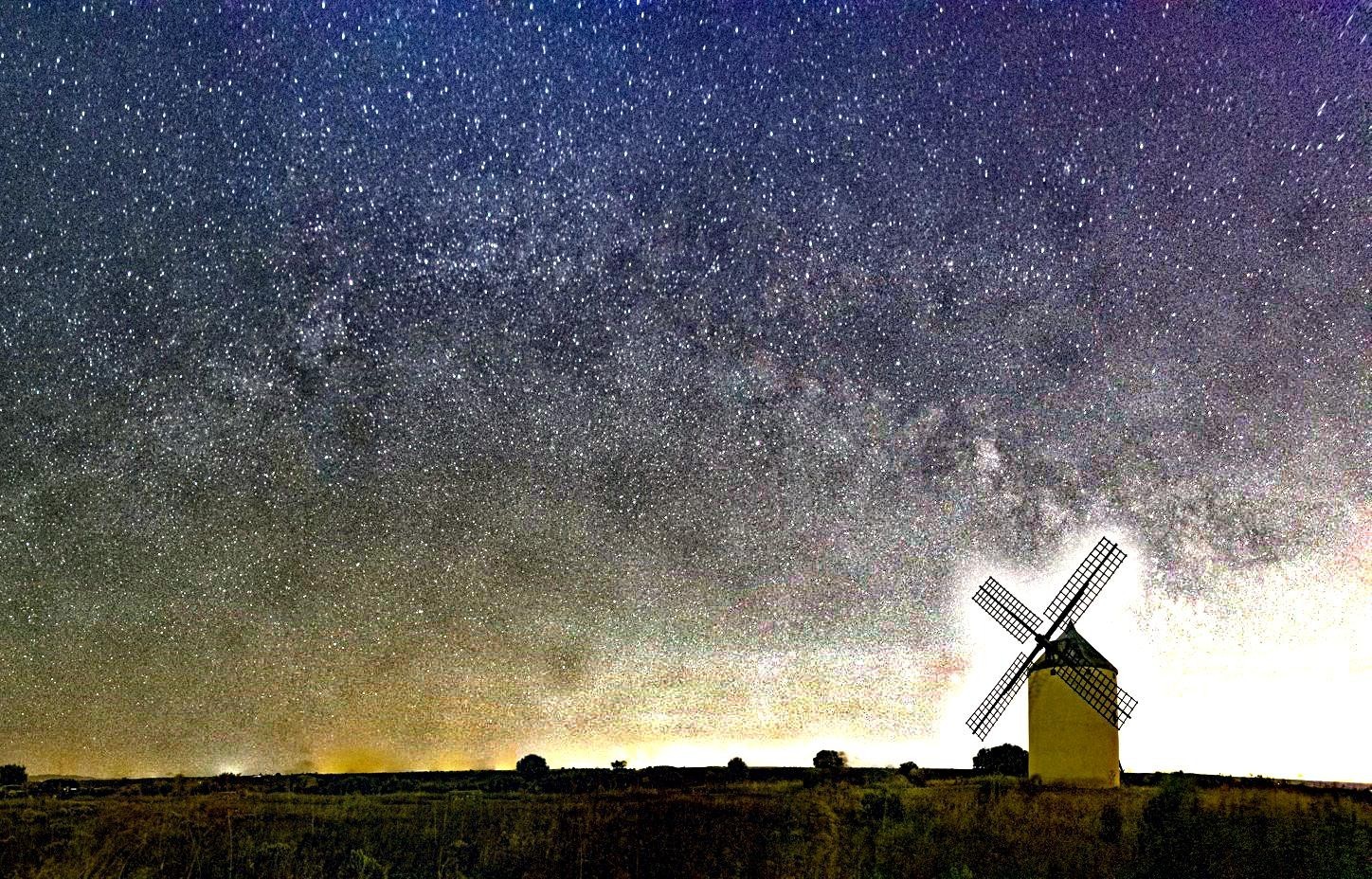}
      \caption{Nighttime dehazing}
    \end{subfigure}

    \begin{subfigure}{0.32\linewidth}
      \centering
      \includegraphics[width=\linewidth,height=0.48\linewidth]{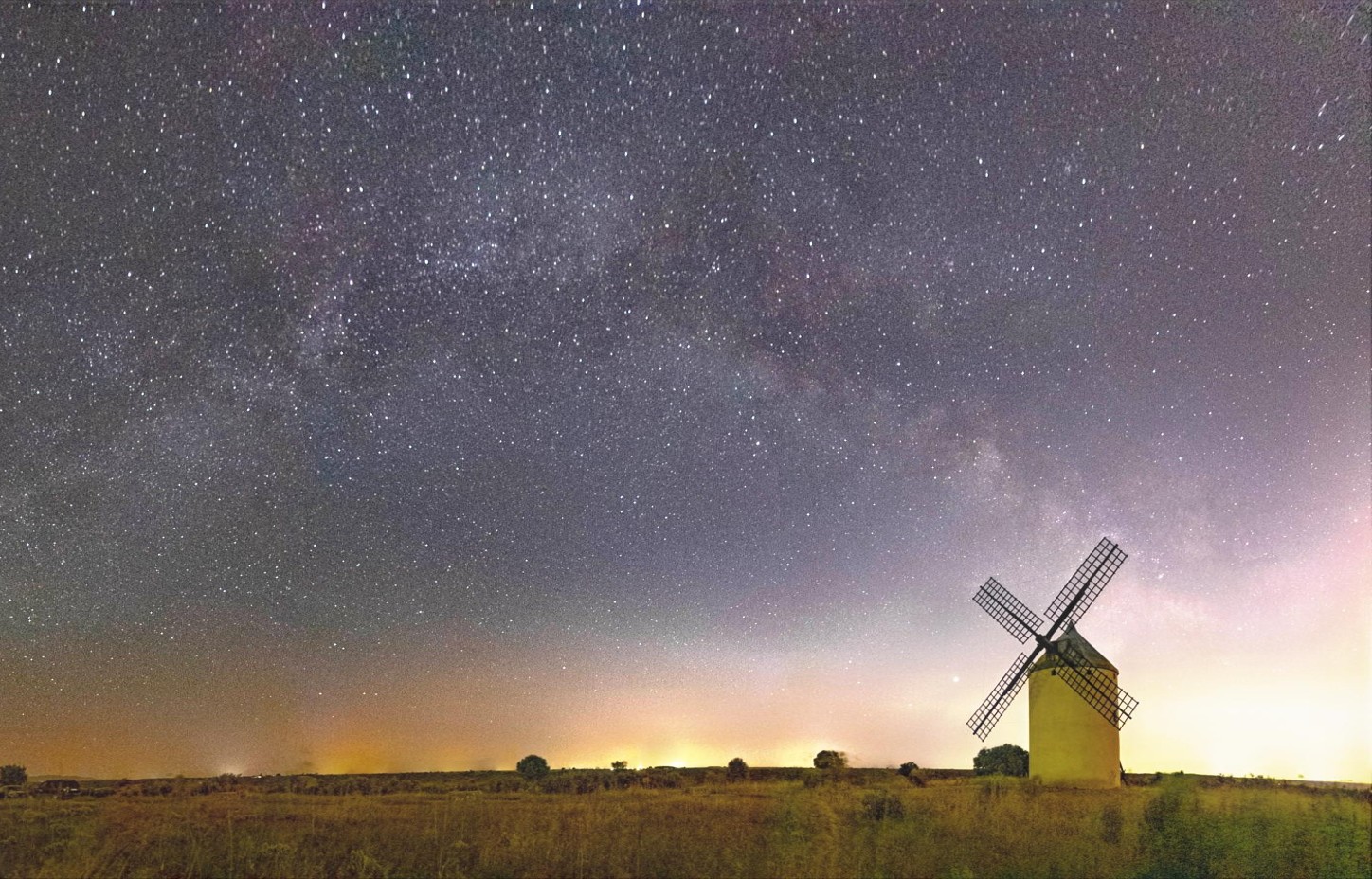}
      \caption{DCPDN}
    \end{subfigure}
    \begin{subfigure}{0.32\linewidth}
      \centering
      \includegraphics[width=\linewidth,height=0.48\linewidth]{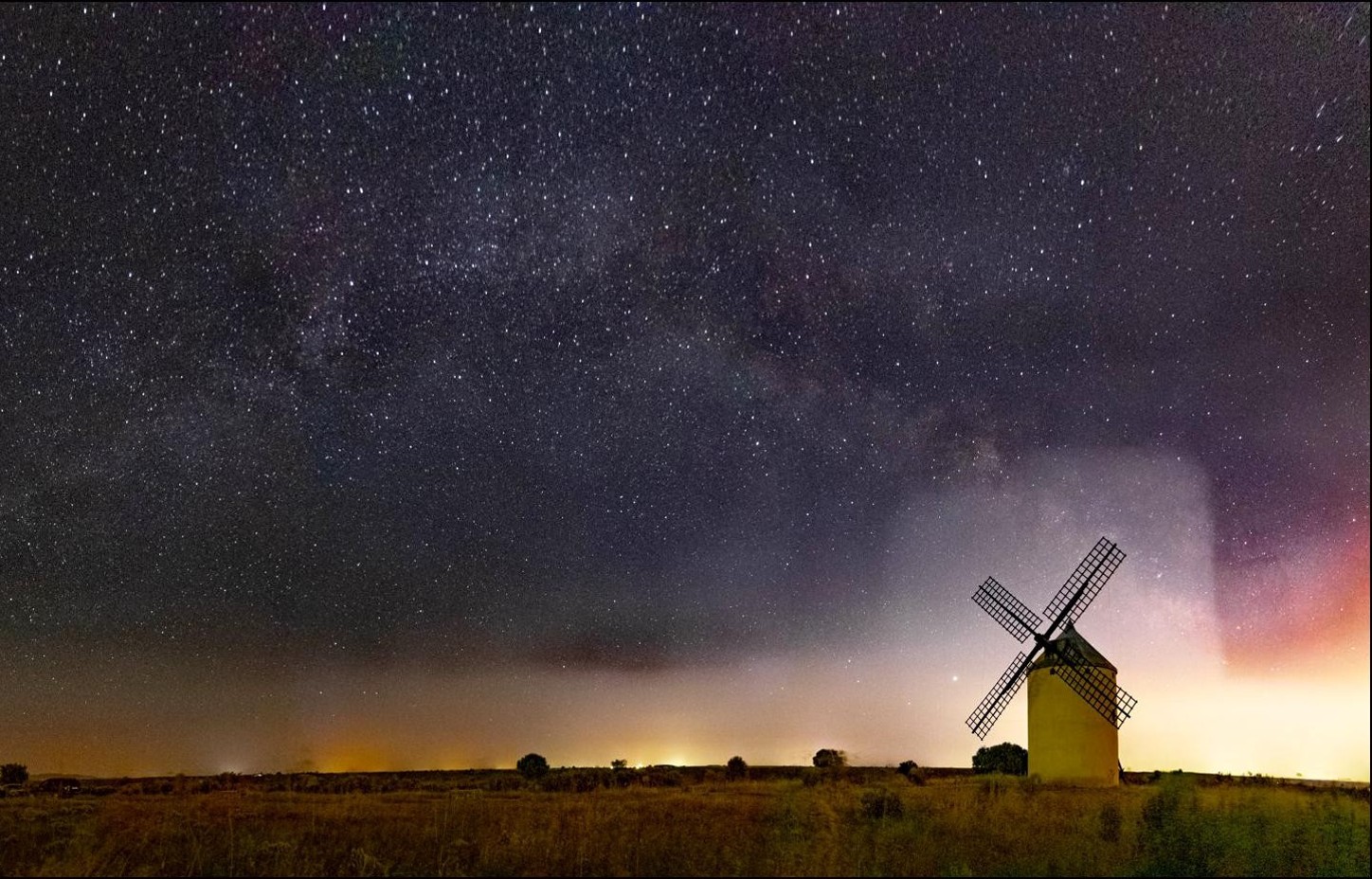}
      \caption{GridDehazeNet}
    \end{subfigure}
     \begin{subfigure}{0.32\linewidth}
       \centering
       \includegraphics[width=\linewidth,height=0.48\linewidth]{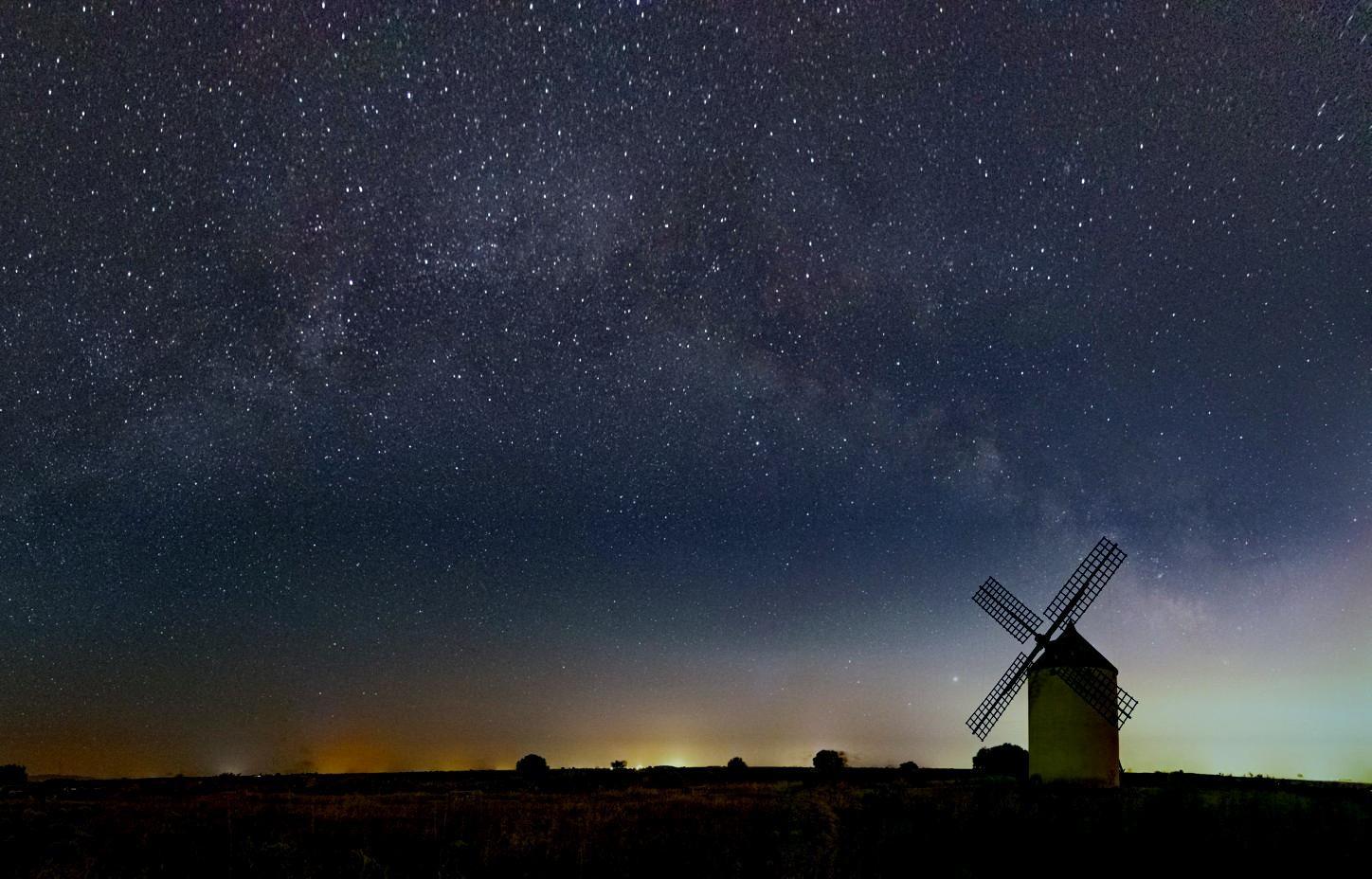}
       \caption{LPR}
     \end{subfigure}
   \caption{Results of the eight tested methods on a nighttime light-polluted sky image.}
 \label{fig:night_sky1}
 \end{figure*}

 \begin{figure*}[t]
   \centering
     \begin{subfigure}{0.32\linewidth}
       \centering
       \includegraphics[width=\linewidth,height=0.48\linewidth]{figures/results/crop/3495.jpg}
       \caption{Light-polluted image}
     \end{subfigure}
     \begin{subfigure}{0.32\linewidth}
       \centering
       \includegraphics[width=\linewidth,height=0.48\linewidth]{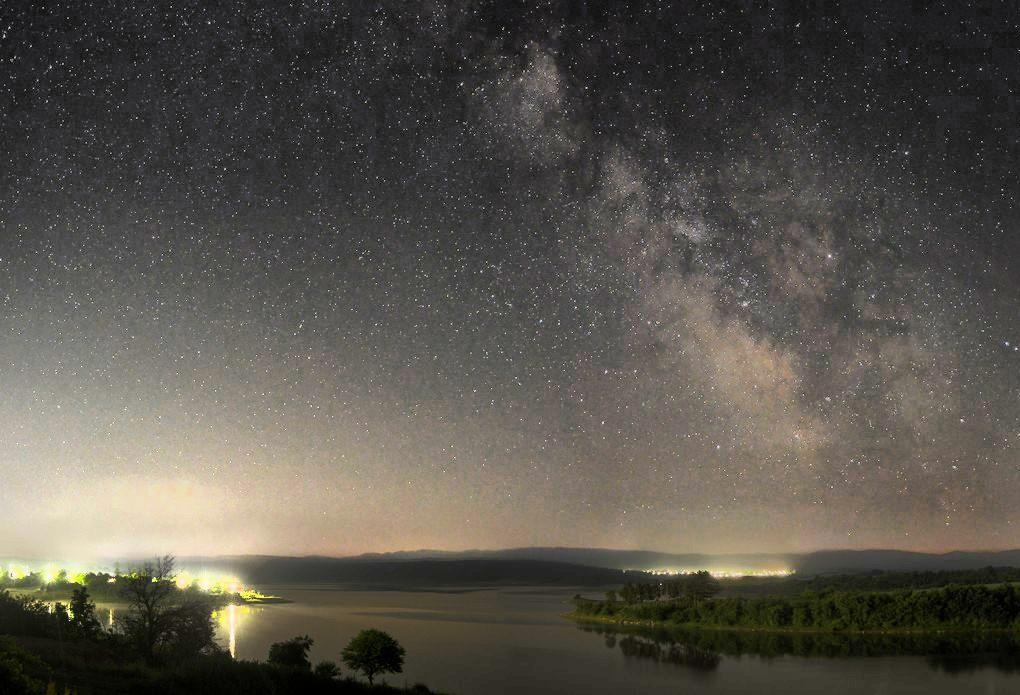}
       \caption{CLAHE}
     \end{subfigure}
     \begin{subfigure}{0.32\linewidth}
       \centering
       \includegraphics[width=\linewidth,height=0.48\linewidth]{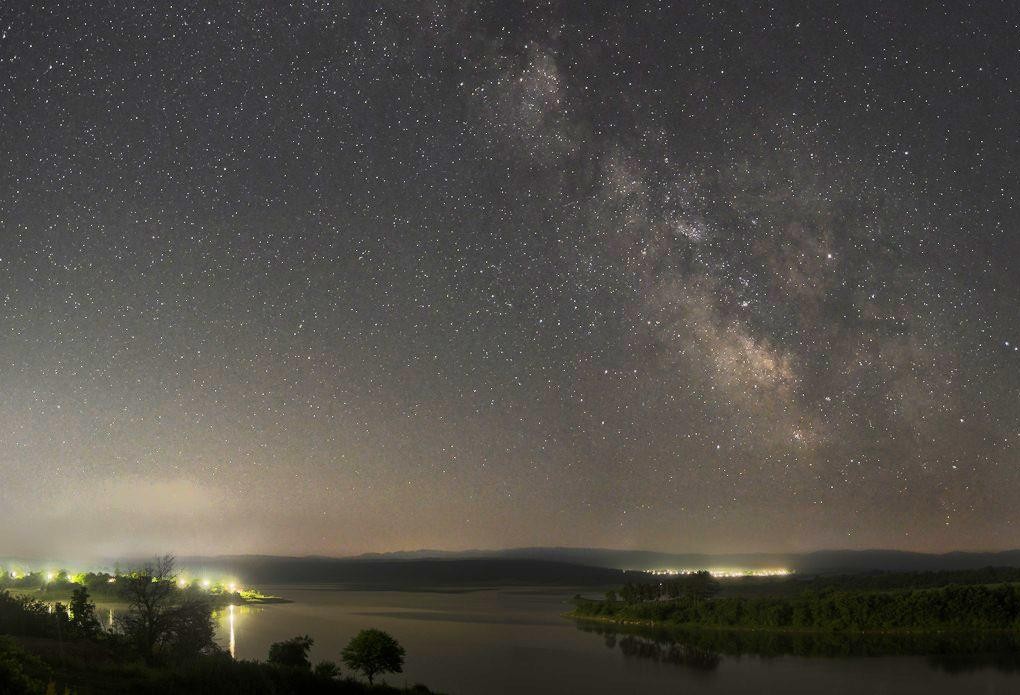}
       \caption{OCTM}
     \end{subfigure}

     \begin{subfigure}{0.32\linewidth}
       \centering
       \includegraphics[width=\linewidth,height=0.48\linewidth]{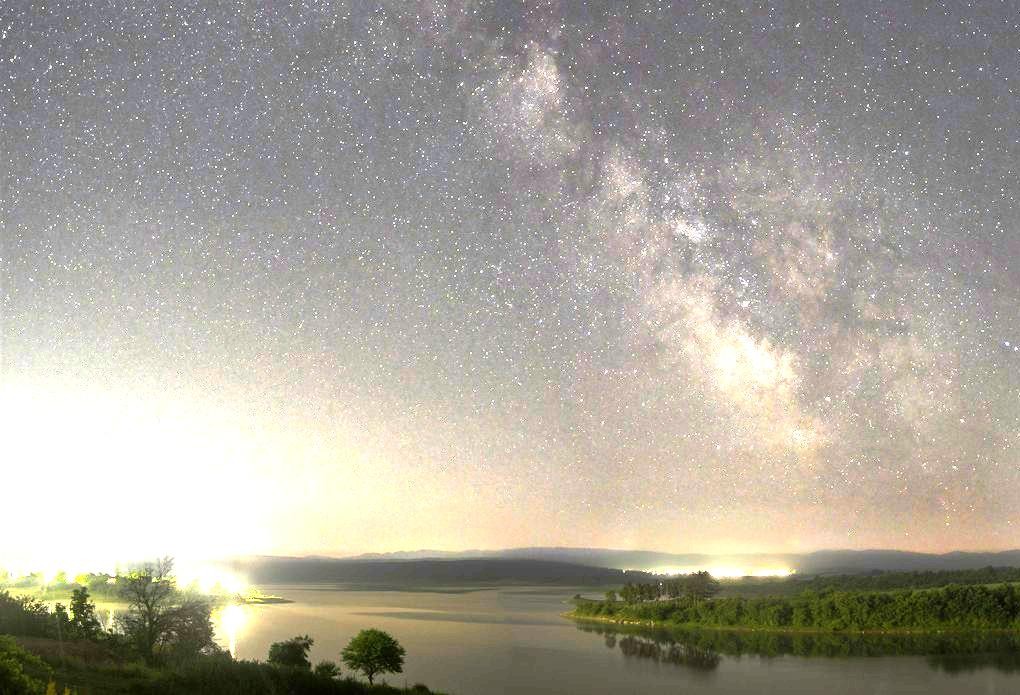}
       \caption{LIME}
     \end{subfigure}
     \begin{subfigure}{0.32\linewidth}
       \centering
       \includegraphics[width=\linewidth,height=0.48\linewidth]{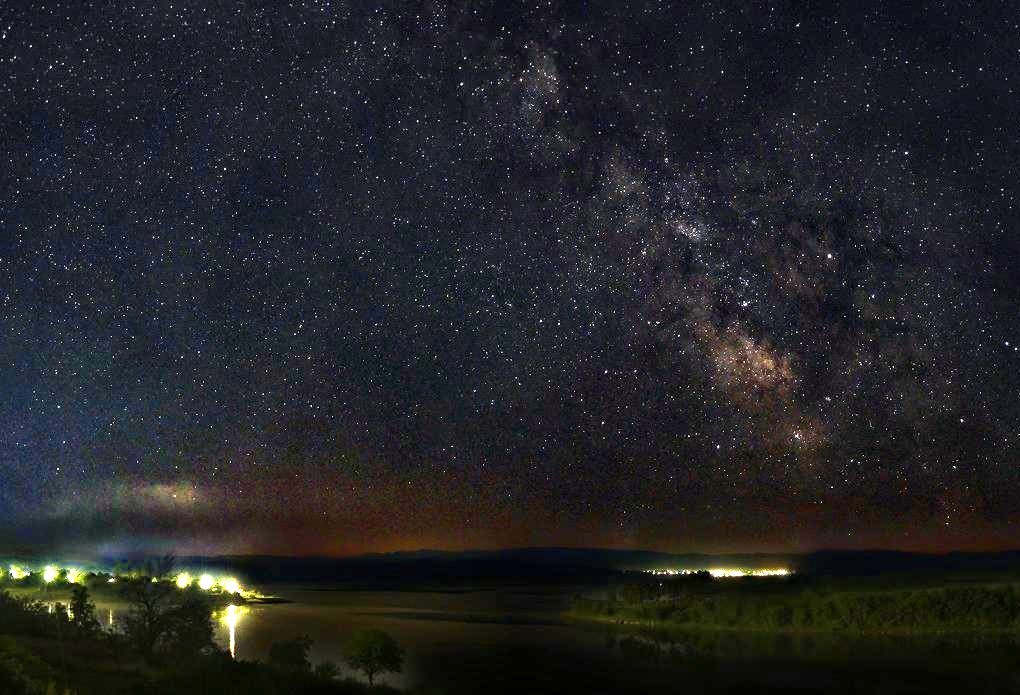}
       \caption{Dark channel dehazing}
     \end{subfigure}
     \begin{subfigure}{0.32\linewidth}
      \centering
      \includegraphics[width=\linewidth,height=0.48\linewidth]{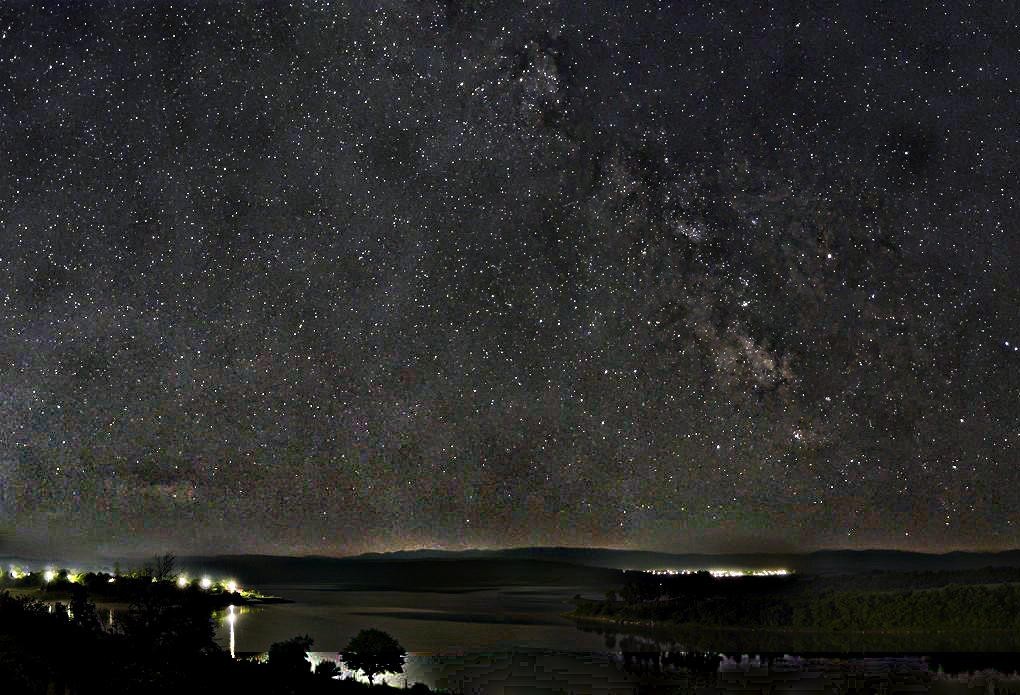}
      \caption{Nighttime dehazing}
    \end{subfigure}

    \begin{subfigure}{0.32\linewidth}
      \centering
      \includegraphics[width=\linewidth,height=0.48\linewidth]{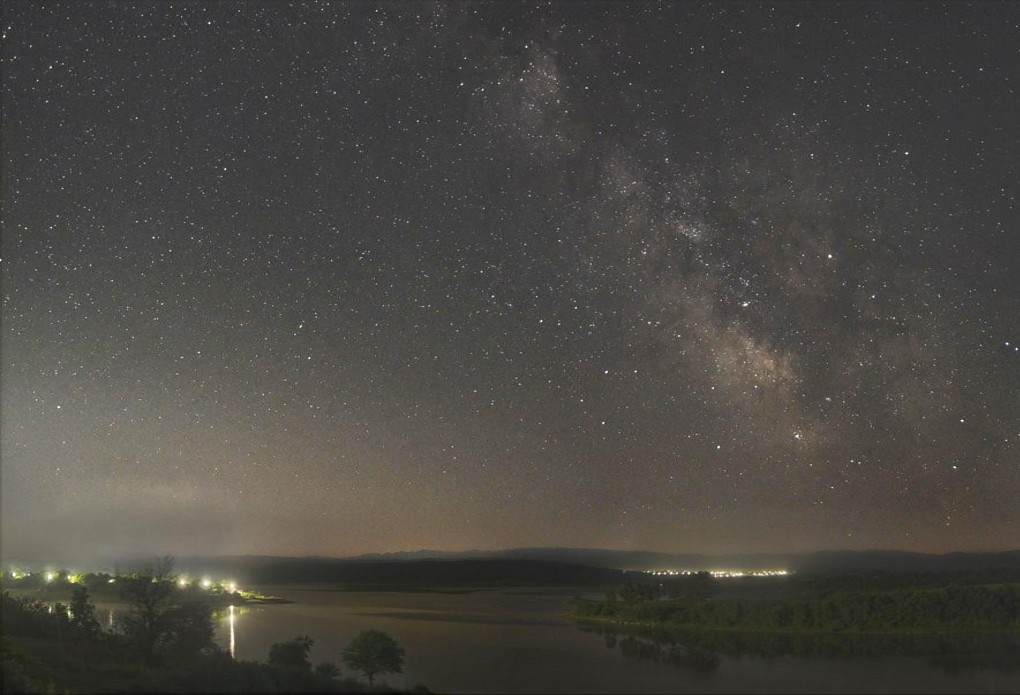}
      \caption{DCPDN}
    \end{subfigure}
    \begin{subfigure}{0.32\linewidth}
      \centering
      \includegraphics[width=\linewidth,height=0.48\linewidth]{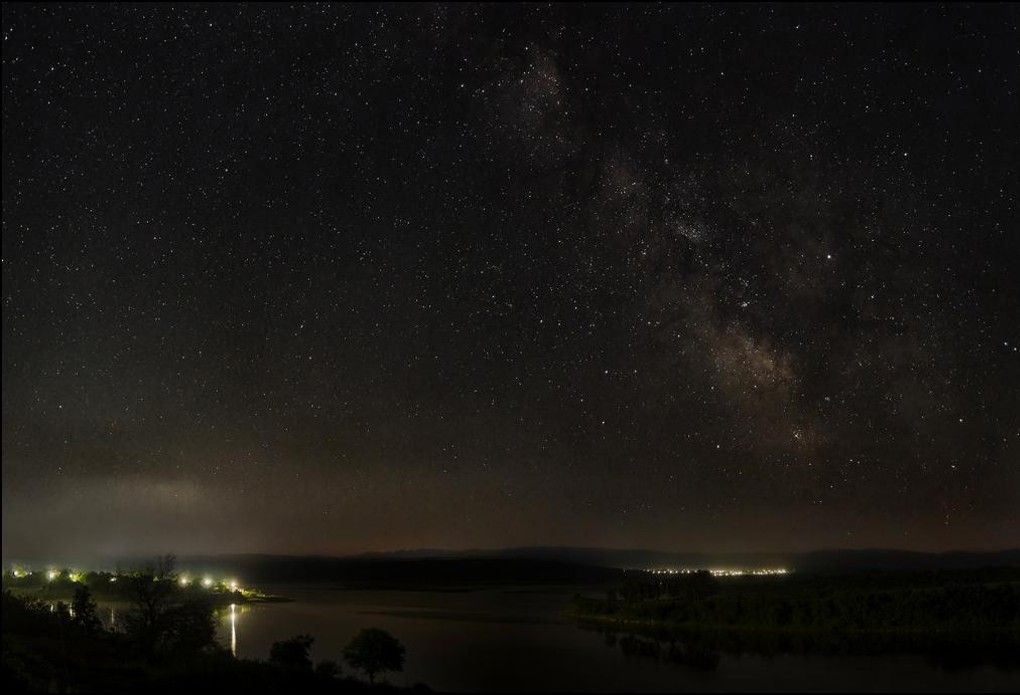}
      \caption{GridDehazeNet}
    \end{subfigure}
     \begin{subfigure}{0.32\linewidth}
       \centering
       \includegraphics[width=\linewidth,height=0.48\linewidth]{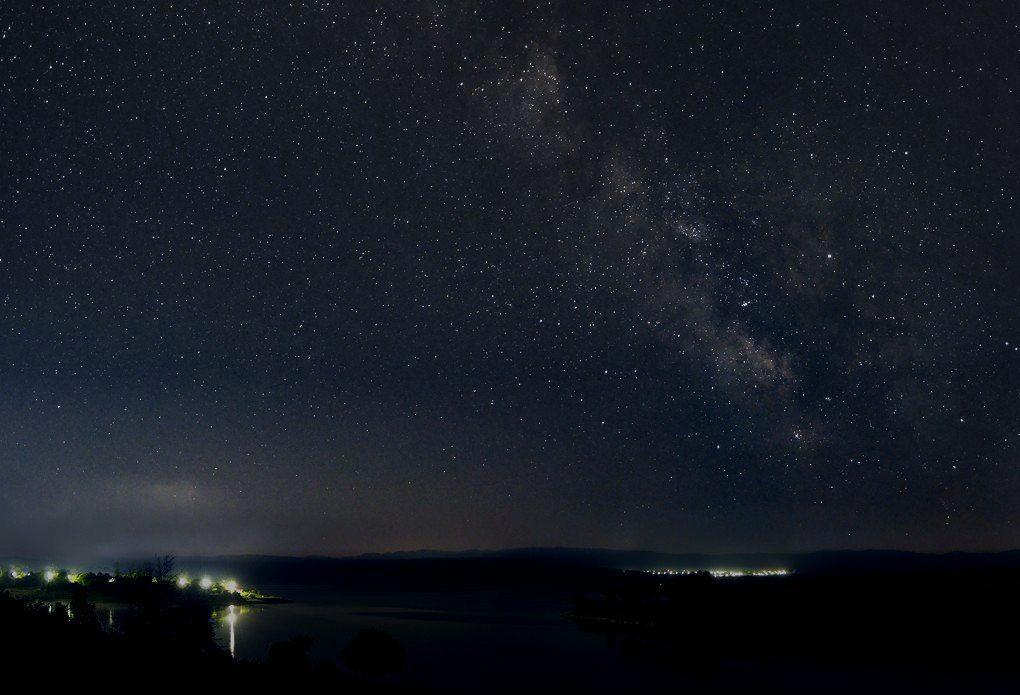}
       \caption{LPR}
     \end{subfigure}
   \caption{Results of the eight tested methods on another nighttime light-polluted sky image.}
 \label{fig:night_sky2}
 \end{figure*}

 \begin{figure*}[t]
   \centering
     \begin{subfigure}{0.46\linewidth}
       \centering
       \includegraphics[width=\linewidth,height=0.45\linewidth]{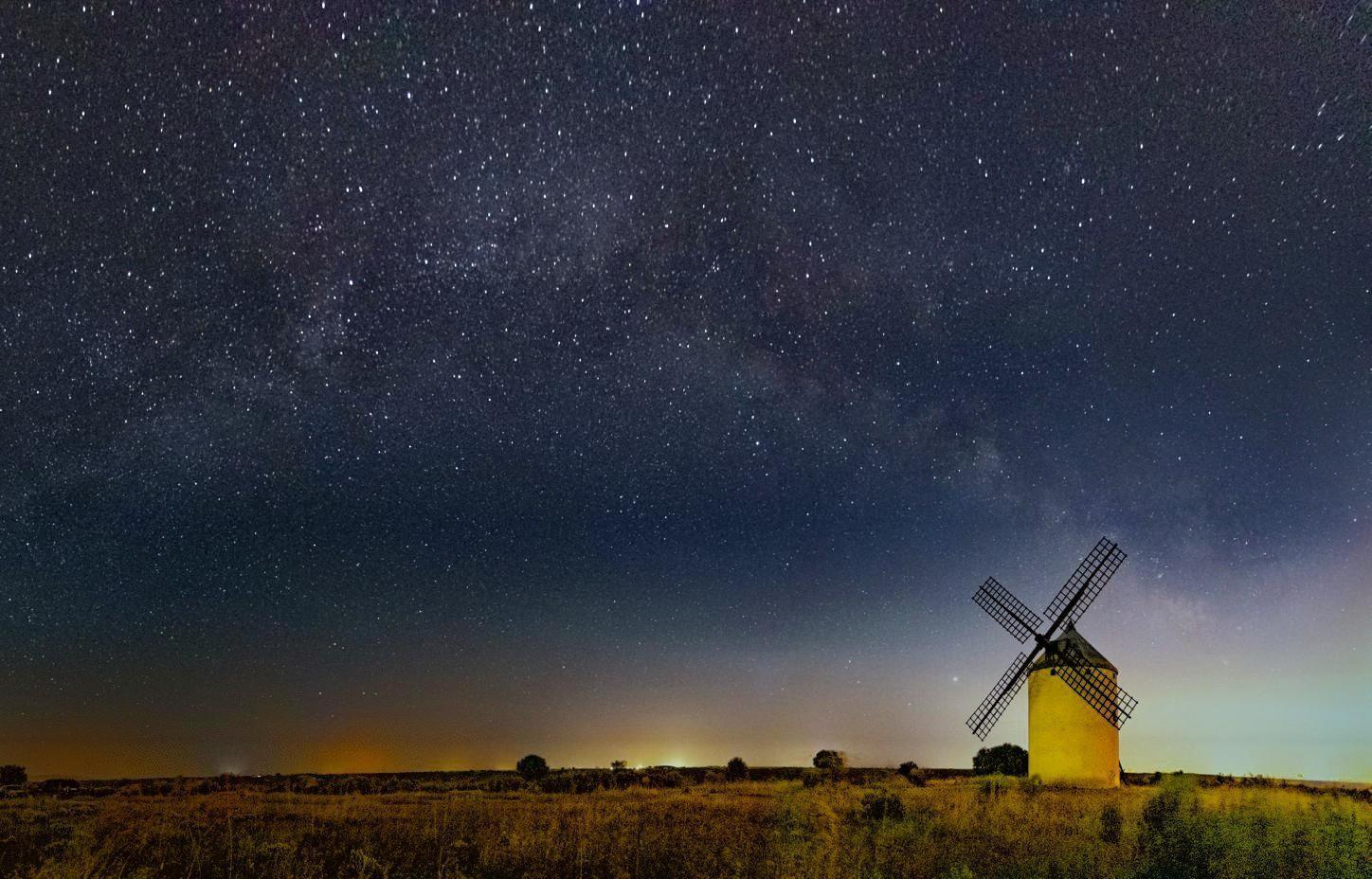}
     \end{subfigure}
     \begin{subfigure}{0.46\linewidth}
       \centering
       \includegraphics[width=\linewidth,height=0.45\linewidth]{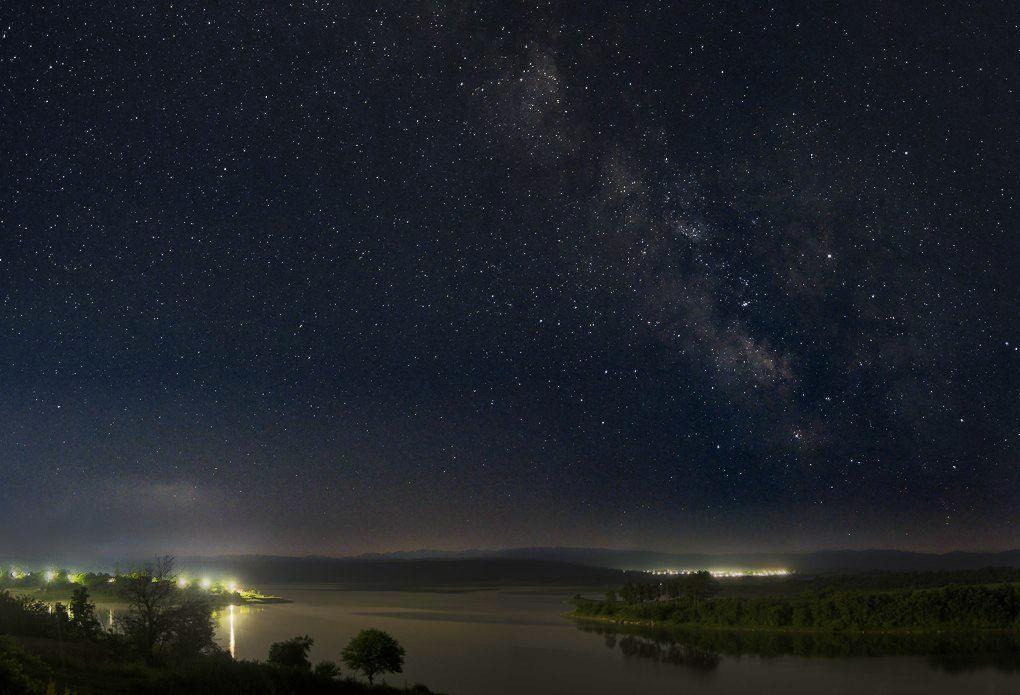}
     \end{subfigure}
    \caption{LPR-OCTM fusion results to be compared with Fig.~\ref{fig:night_sky1}(i) and Fig.~\ref{fig:night_sky2}(i).}
 \label{fig:hybrid}
 \end{figure*}

 \begin{figure*}[!htbp]
   \centering
     \begin{subfigure}{0.32\linewidth}
       \centering
       \includegraphics[width=\linewidth,height=0.48\linewidth]{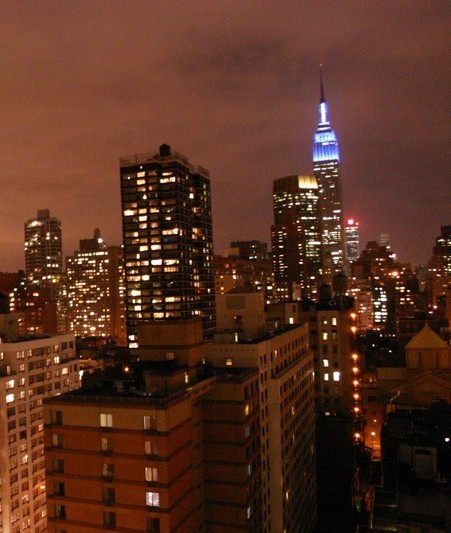}
       \caption{Light-polluted image}
     \end{subfigure}
     \begin{subfigure}{0.32\linewidth}
       \centering
       \includegraphics[width=\linewidth,height=0.48\linewidth]{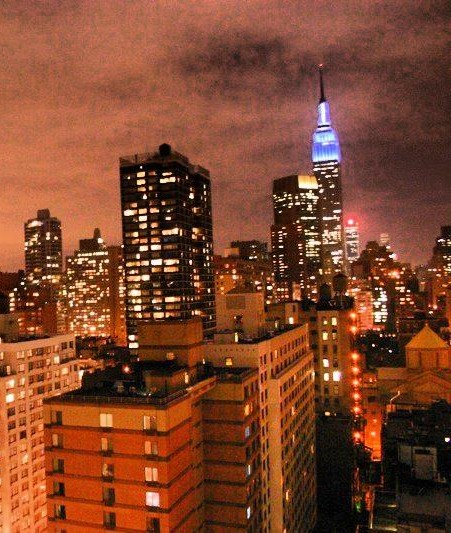}
       \caption{CLAHE}
     \end{subfigure}
     \begin{subfigure}{0.32\linewidth}
       \centering
       \includegraphics[width=\linewidth,height=0.48\linewidth]{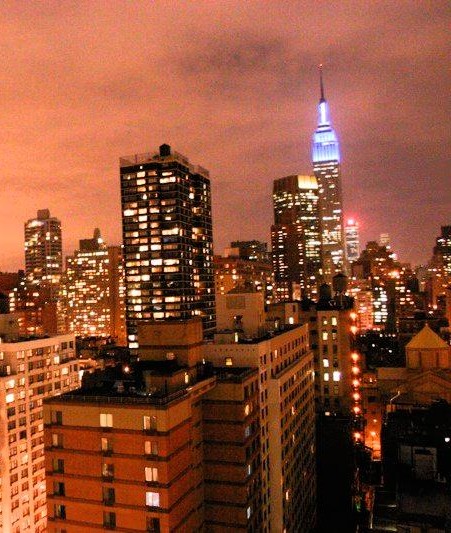}
       \caption{OCTM}
     \end{subfigure}

     \begin{subfigure}{0.32\linewidth}
       \centering
       \includegraphics[width=\linewidth,height=0.48\linewidth]{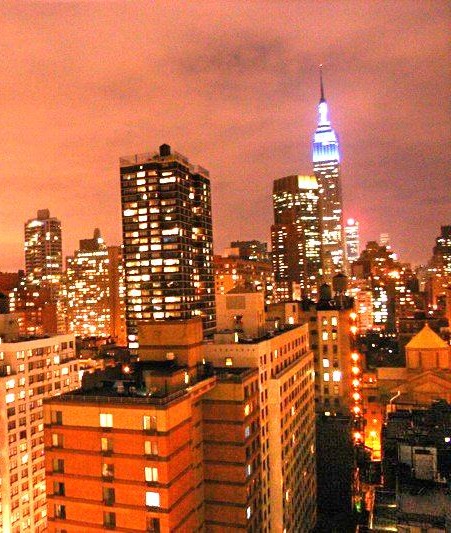}
       \caption{LIME}
     \end{subfigure}
     \begin{subfigure}{0.32\linewidth}
       \centering
       \includegraphics[width=\linewidth,height=0.48\linewidth]{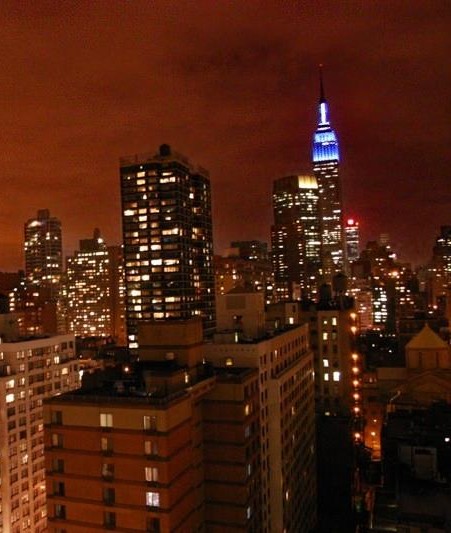}
       \caption{Dark channel dehazing}
     \end{subfigure}
     \begin{subfigure}{0.32\linewidth}
      \centering
      \includegraphics[width=\linewidth,height=0.48\linewidth]{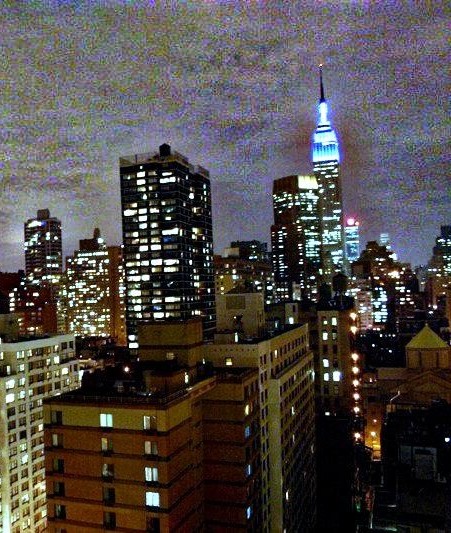}
      \caption{Nighttime dehazing}
    \end{subfigure}

      \begin{subfigure}{0.32\linewidth}
       \centering
       \includegraphics[width=\linewidth,height=0.48\linewidth]{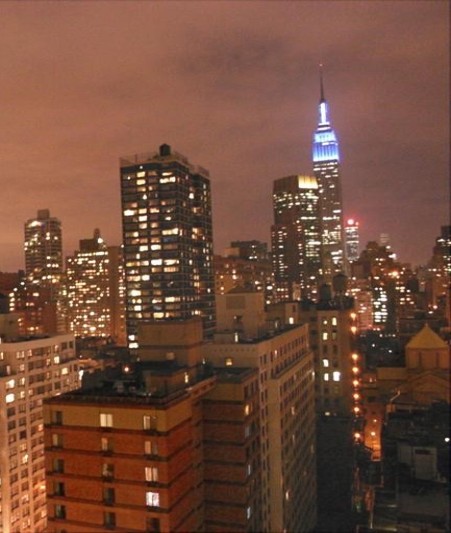}
       \caption{DCPDN}
     \end{subfigure}
     \begin{subfigure}{0.32\linewidth}
      \centering
      \includegraphics[width=\linewidth,height=0.48\linewidth]{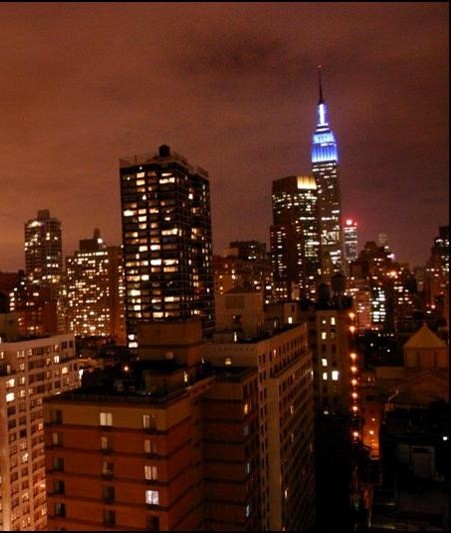}
      \caption{GridDehazeNet}
    \end{subfigure}
     \begin{subfigure}{0.32\linewidth}
       \centering
       \includegraphics[width=\linewidth,height=0.48\linewidth]{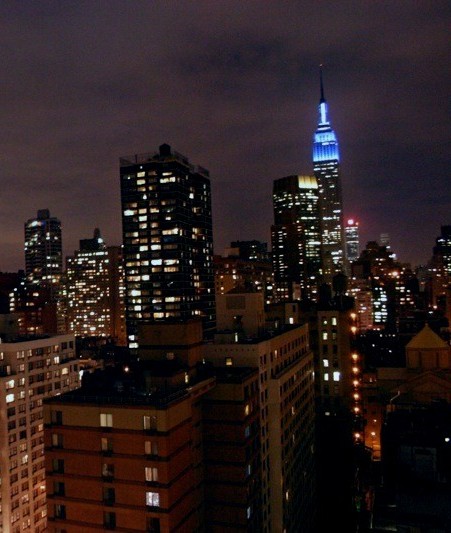}
       \caption{LPR}
     \end{subfigure}
   \caption{Results of the eight tested methods on another nighttime light-polluted city image.}
 \label{fig:night_city2}
 \end{figure*}

 \begin{figure*}[t]
   \centering
   \setlength{\abovecaptionskip}{4pt}
     \begin{subfigure}{0.32\linewidth}
       \centering
       \includegraphics[width=\linewidth,height=0.53\linewidth]{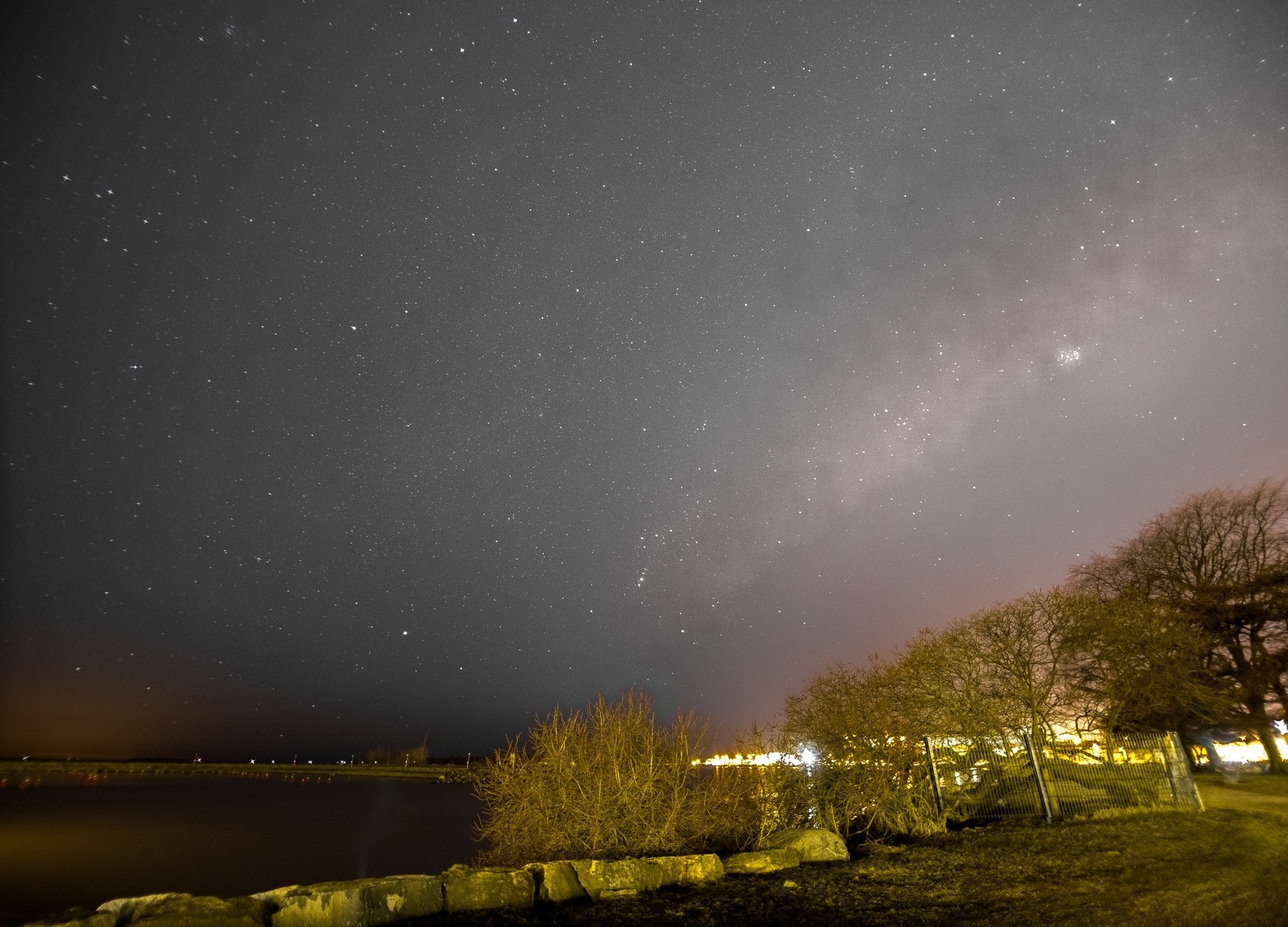}
       \caption{Light-polluted image}
     \end{subfigure}
     \begin{subfigure}{0.32\linewidth}
       \centering
       \includegraphics[width=\linewidth,height=0.53\linewidth]{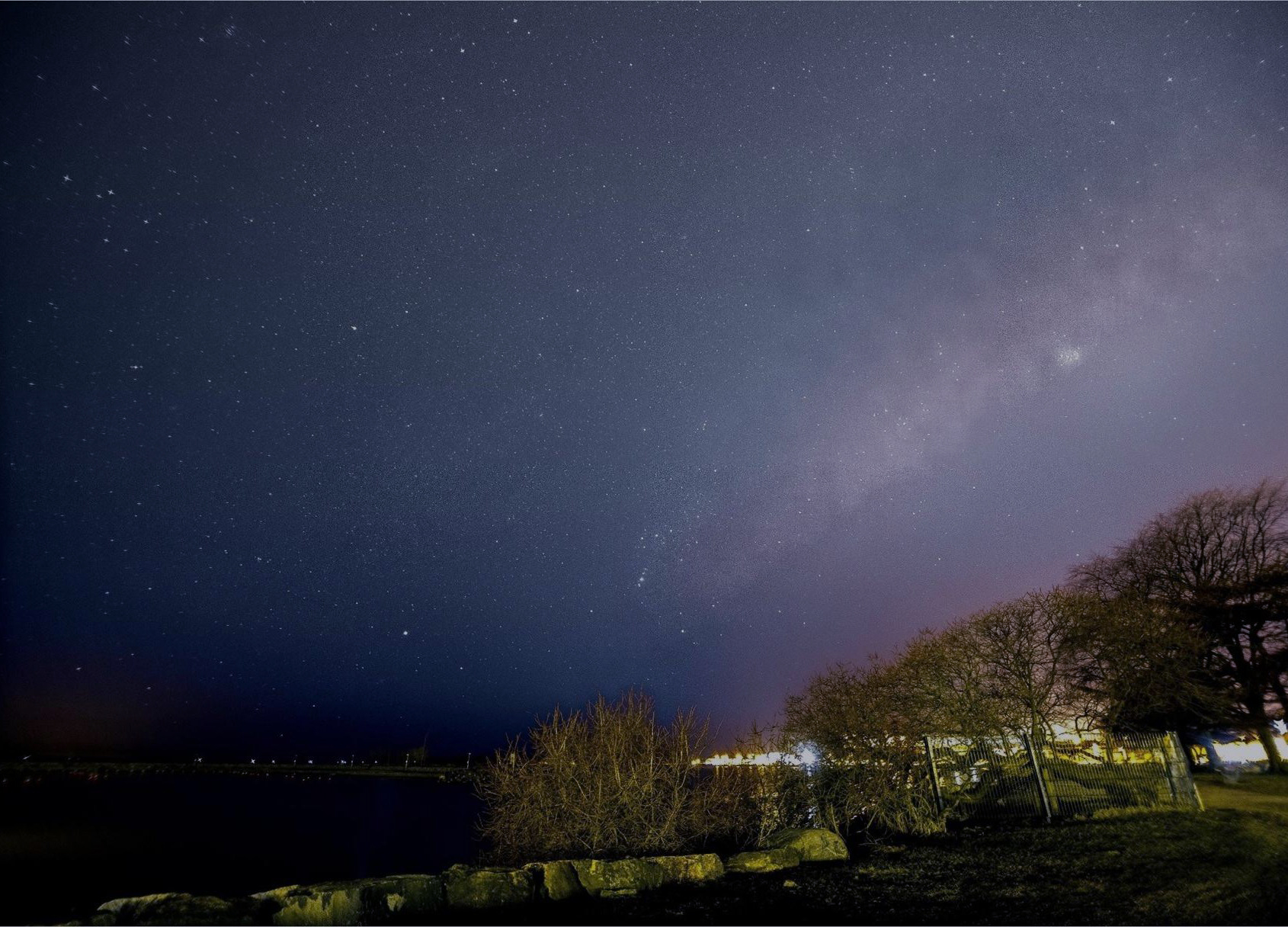}
       \caption{Baseline method}
     \end{subfigure}
     \begin{subfigure}{0.32\linewidth}
       \centering
       \includegraphics[width=\linewidth,height=0.53\linewidth]{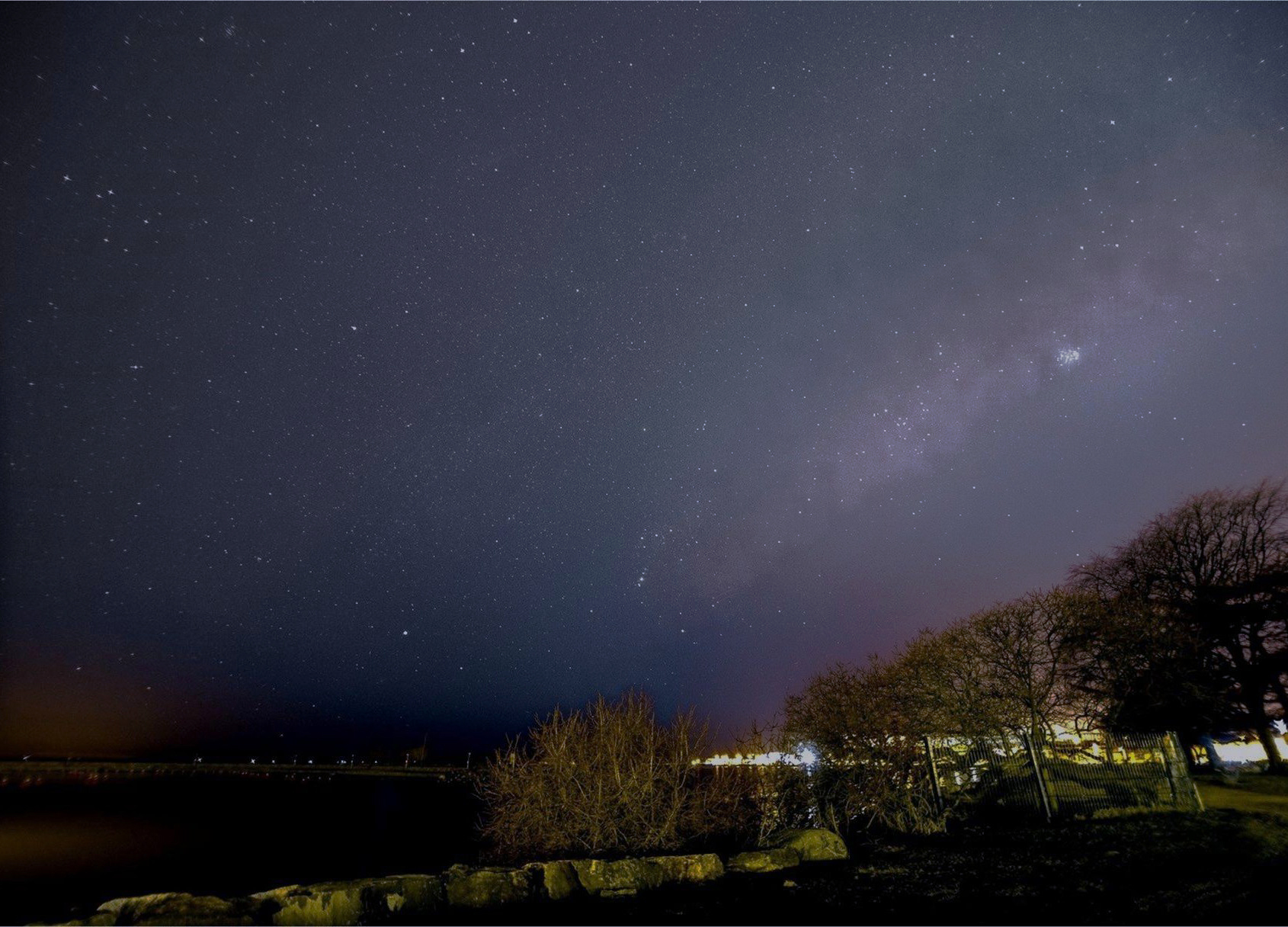}
       \caption{Adaptive method}
     \end{subfigure}
   \caption{Light pollution reduction results by the baseline and adaptive methods.}
 \label{fig:baseline_adaptive}
 \vspace{-0.4cm}
 \end{figure*}

 \begin{figure*}[t]
   \centering
   \setlength{\abovecaptionskip}{4pt}
     \begin{subfigure}{0.32\linewidth}
       \centering
       \includegraphics[width=\linewidth]{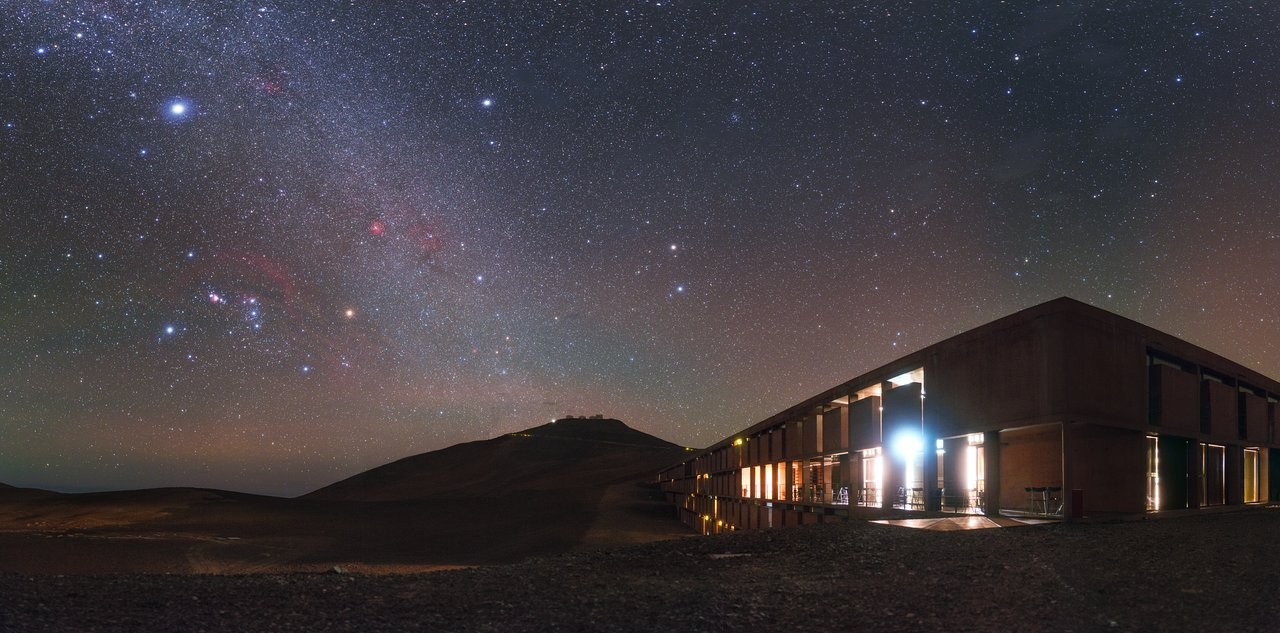}
       \caption{Light-polluted image}
     \end{subfigure}
     \begin{subfigure}{0.32\linewidth}
       \centering
       \includegraphics[width=\linewidth]{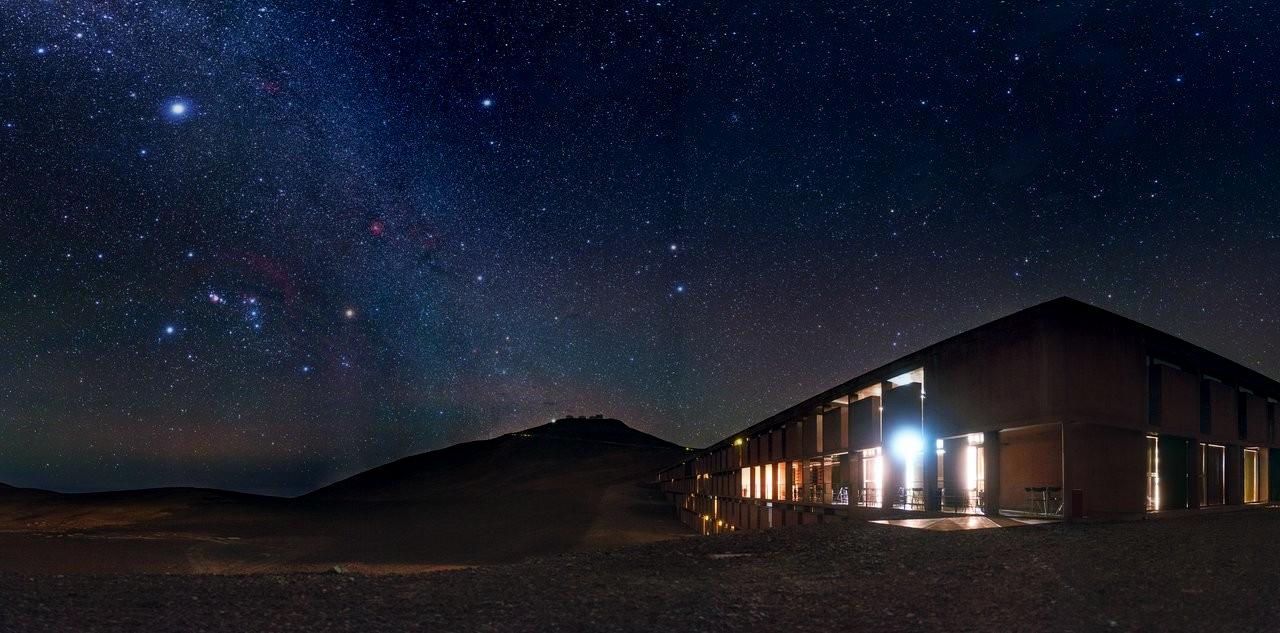}
       \caption{LPR ($\beta=10^{-4}$)}
     \end{subfigure}
     \begin{subfigure}{0.32\linewidth}
       \centering
       \includegraphics[width=\linewidth]{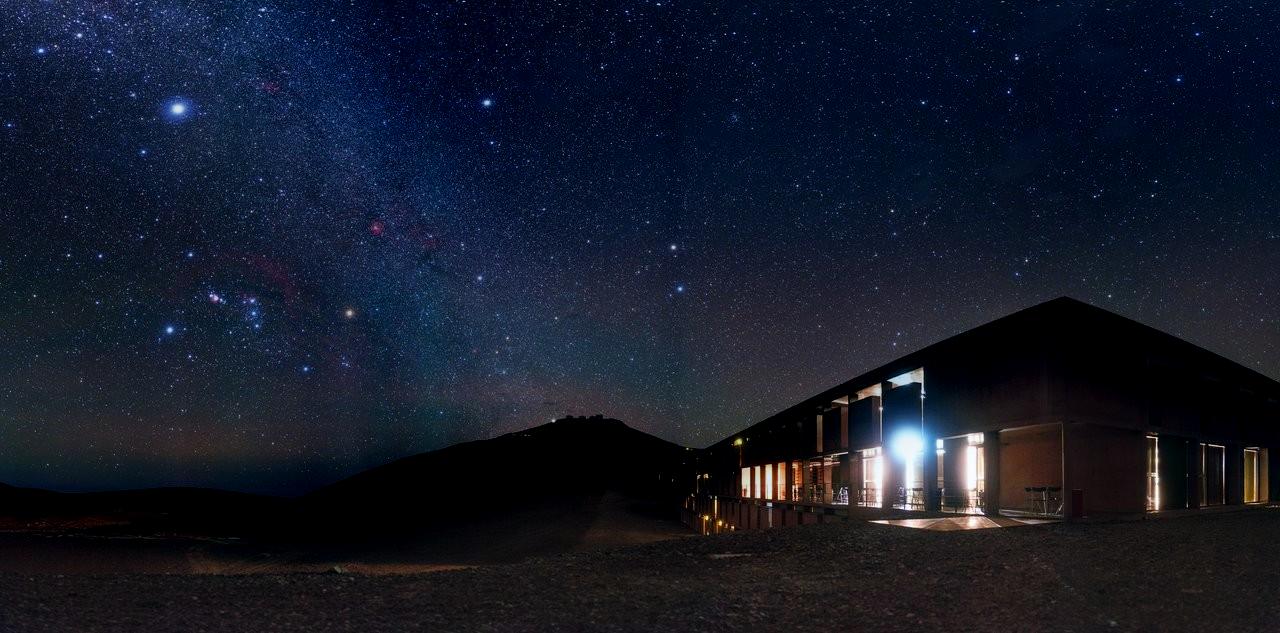}
       \caption{LPR ($\beta=10^{-3}$)}
     \end{subfigure}
   \caption{Light pollution reduction results with different scattering coefficients.}
 \label{fig:beta}
 \vspace{-0.4cm}
 \end{figure*}



 Besides the sky, unwanted artificial lights can also contaminate the parts of a nighttime image below skyline.  For example, when shooting elaborately illuminated structures such as bridges or buildings in distance, in the light pathway between the target objects and camera there are other artificial lights.  These in-between lights can reduce the contrast and dynamic range, and also distort the color of the intended image depending on the spectral signatures of artificial lights.  Our LPR method can be applied, if combined with some depth information, to restore light-polluted urban landscape images.

 When applying Eq(\ref{eq:J(x,y)_ada}) to restore pixels below skyline, we need to set the integral upper limit $L$ to the distance of the imaged object to the camera, rather than simply letting $L = \infty$ for those pixels above the skyline.  The required depth information can be obtained by one of several single-image depth estimation algorithms \cite{godard2019digging,roy2016monocular,xian2020structure}.
 Unlike in computer vision tasks, the goal of LPR is perceptual image quality and hence it does not need very high precision in estimated depth values.  For our task the most important information is the depth rank.  Thanks to recent progresses of deep learning in computer vision, many of existing depth estimation algorithms offer acceptable precision for the purpose of light pollution reduction.

 But single-image depth estimation algorithms have a common shortcoming that negatively affects perceptual image quality, if their depth results are directly fed into our restoration algorithm.  This shortcoming is relatively low spatial resolution in depth discontinuity (blurred depth edges).  The poor spatial resolution of depth map can cause
 halos and blurs in restored nighttime urban landscape images as shown in Fig.~\ref{fig:halos}(c).  We rectify the problem by edge-guided filtering \cite{he2012guided} of the estimated depth map.  This enforces the alignment of the depth edges and corresponding edges in input image $\hat{I}(x,y)$.  Fig.~\ref{fig:halos} demonstrates how
 edge-guided filtering improves perceptual quality of the restored image.

\section{Experiments}

In this section, we present and evaluate the experimental results of the proposed LPR algorithm.  All the test images are found in  the internet with keyword "light pollution".  For all light-polluted test images there are no corresponding pollution-free ground truth images, hence the evaluations are necessarily based on subjective image quality.  Also, as the LPR algorithm is the first of its kind, comparison studies can only be carried out against other image enhancement and tone mapping algorithms that are not specifically designed for light pollution removal but can be used to increase contrast and dynamic range of nighttime images.  These competing algorithms include the contrast limited adaptive histogram equalization (CLAHE) \cite{zuiderveld1994contrast}, optimal contrast-tone mapping (OCTM) \cite{wu2010linear}, and the LIME method for enhancing low light images \cite{guo2016lime}.  

Also, we add four dehazing methods in the comparison group, including the dark channel dehazing algorithm \cite{he2010single}, a nighttime dehazing algorithm \cite{li2015nighttime}, and two recently published deep learning based dehazing algorithms DCPDN \cite{zhang2018densely} and GridDehazeNet\cite{liu2019griddehazenet}.
This is because dehazing is a similar image restoration task, namely, removing unwanted effects of light scattering by aerosols.

\subsection{Nighttime natural landscapes}

In Fig.~\ref{fig:title} we have seen clearly how the LPR algorithm removes light pollution in the atmosphere and restores night skies closer to the brightness and color in absence of artificial lights.  Figs.~\ref{fig:night_sky1} and \ref{fig:night_sky2} present the results of the eight methods in the comparison group on two more sample images of nighttime natural landscapes.  The LPR algorithm is a clear winner among the eight methods in terms of restoring the night sky in a state free of artificial light pollution.  LPR greatly improves the visual appeal of night sky images by reducing the background brightness above the horizon and enhancing the stars and cloud textures.  The other seven methods also enhance the stars and clouds in the sky but they suffer from various color distortions and other artifacts.  The enhancement methods CLACH and OCTM adjust the sky brightness in opposite way, increasing instead of decreasing it.  The LIME method turns the input night image into a daylight image. The dark channel method does dim the night sky in compensation for the scattering of artificial lights in atmosphere.  But it causes severe objectionable color shifts and halo artifacts (see the windmill contour in Fig.~\ref{fig:night_sky1}(e)). The nighttime dehazing method enhances the contrast but produces severe artifacts in the sky region.  The deep learning based dehazing methods DCPDN increases the overall brightness and reduces the color saturation. The GridDehazeNet generates false contours above the skyline (see  Fig.~\ref{fig:night_sky1}(h)).

In Fig.~\ref{fig:night_sky2}(c), the OCTM algorithm enhances the lake and woods below the skyline without increasing the brightness of sky too much as in the CLACH and LIME methods.  Depending on personal preference, some viewers may like the OCTM effects below the skyline.  This suggests a way to combine the best parts of LPR and OCTM, and merge the results of the two algorithms along the skyline into a more balanced and visually even more pleasing final output image.  We present, in Fig.~\ref{fig:hybrid}, such LPR-OCTM fusion results of the two nighttime landscape images in Figs.~\ref{fig:night_sky1} and \ref{fig:night_sky2}.

\subsection{Nighttime urban scenes}

Fig.~\ref{fig:night_city2} presents the results of the eight different methods on one nighttime downtown images of heavy light pollution.  CLACH, OCTM and LIME methods fail these challenging tests badly.
The dark channel method reduces the overall brightness somewhat and increases contrast modestly.  But like when being used in the task of removing light pollution in nighttime natural scenes, the dehazing method generates color shifts.
For the other three dehazing methods, similar conclusions can be made as in the case of nighttime natural scenes.
Only the LPR algorithm passes the tests and successfully removes much of light pollution.  It dims the sky, noticeably increases the overall dynamic range, and enhances surface details of the buildings.



\subsection{Ablation study}

All the above results are generated by the spatially adaptive version of the LPR algorithm.  Fig.~\ref{fig:baseline_adaptive} lets the reader visually examine what changes will take place if the baseline version of the LPR algorithm is used.  In this test image, the radiance of artificial lights is not uniformly distributed on the ground; the pollution radiance is much higher on the right side of the image than the left side.  The oversimplified $x$-invariant pollution radiance model of Eq (\ref{eq:Ep(x)}) is clearly inaccurate. Therefore, the baseline LPR algorithm cannot compensate for the spatial variations of the pollution light radiance.  This causes the upper sky portion of the restored image Fig.~\ref{fig:baseline_adaptive}(b) to have an increasing intensity ramp from left to right, i.e., still exhibiting a pattern correlated to artificial lights.

Finally, we discuss how to make tradeoffs between different perceptual goals by setting air quality parameter $\beta$ in the LPR algorithm.  Fig.~\ref{fig:beta} compares the LPR results on a test image for assuming air is relatively clean and transparent ($\beta=10^{-4}$) vs.\ less so ($\beta=10^{-3}$).  The larger the value of $\beta$ (the higher density of aerosols in the atmosphere), the more scattered light energy is removed from the input image by the LPR algorithm. Consequently, the residual effects of artificial lights become lesser, but some subtle details revealed by natural lights via long-exposure photography may get suppressed.  Note the disappeared mountain top silhouette from (b) to (c) in ~\ref{fig:beta}.
All experimental results reported above are generated with $\beta=10^{-4}$.

\section{Conclusions}

We designed, implemented and experimented with a light pollution reduction algorithm for the task of alleviating adverse visual effects of unwanted artificial lights in nighttime photography.  The algorithm is derived from a physical image formation model that accounts for the interactions of artificial lights, aerosols in atmosphere and the camera; it can characterize pollution light sources and to a large degree neutralize them in restored nighttime images of both nature and urban landscapes.

{\small
\bibliographystyle{ieee_fullname}
\bibliography{DeLightPollution-bibliography}
}

\end{document}